\documentclass[10pt,journal,compsoc]{IEEEtran}

\ifCLASSOPTIONcompsoc
  \usepackage[nocompress]{cite}
\else
  \usepackage{cite}
\fi
\ifCLASSINFOpdf
\else
\fi
\usepackage{times}
\usepackage{epsfig}
\usepackage{graphicx}
\usepackage{amsmath}
\usepackage{amssymb}
\usepackage{amsfonts}
\usepackage{tabularx}
\usepackage{bbm}

\usepackage{balance}
\usepackage{rotating}
\usepackage{booktabs}
\usepackage[pagebackref=true,breaklinks=true,letterpaper=true,colorlinks,bookmarks=false]{hyperref}

\usepackage{mathtools}
\usepackage{multirow}

\usepackage{graphicx, nicefrac}
\usepackage{subfig}
\usepackage{enumitem}
\usepackage[ruled,linesnumbered]{algorithm2e}
\usepackage{booktabs}
\usepackage{multirow}
\usepackage{xcolor}
\newcommand{\eg}[0]{\emph{e.g., }}
\newcommand{\ie}[0]{\emph{i.e., }}

\makeatletter

\graphicspath{{images/}}

\def\BState{\State\hskip-\ALG@thistlm}

\newcommand{\comment}[1]{}

\newcommand{\mB}[0]{\mathcal{B}}

\newcommand{\mL}[0]{\mathcal{L}}

\newcommand{\mM}[0]{\mathcal{M}}

\newcommand{\mV}[0]{\mathcal{V}}
\newcommand{\mI}[0]{\mathcal{I}}
\newcommand{\mK}[0]{\mathcal{K}}
\newcommand{\mS}[0]{\mathcal{S}}
\newcommand{\mA}[0]{\mathcal{A}}
\newcommand{\mX}[0]{\mathcal{X}}

\newcommand{\vv}[0]{\mathbf{v}}

\newcommand{\R}{\mathbb{R}}

\makeatletter
\def\endthebibliography{%
	\def\@noitemerr{\@latex@warning{Empty `thebibliography' environment}}%
	\endlist
}
\makeatother

\begin{document}
%
\title{Robust Differentiable SVD}
%
%
%
%

\author{Wei~Wang,
	Zheng~Dang,
	Yinlin~Hu,
	Pascal~Fua,~\IEEEmembership{Fellow,~IEEE,}
	and~Mathieu~Salzmann,~\IEEEmembership{Member,~IEEE,}
	\IEEEcompsocitemizethanks{
		\IEEEcompsocthanksitem Wei~Wang, Yinlin~Hu, Pascal~Fua, and~Mathieu~Salzmann are with CVLab, School of Computer and Communication Sciences, EPFL.\protect\\
		E-mail: see https://www.epfl.ch/labs/cvlab/people/ \protect\\
		
		Zheng~Dang is with the National Engineering Laboratory for Visual Information Processing and Application, Xi’an Jiaotong University, Xi’an, China. 
		E-mail: dangzheng713@stu.xjtu.edu.cn\protect\\
		
	}
    
	\thanks{Manuscript received 15 May, 2020; revised 18 January, 2021; accepted 31 March 2021.}
	\thanks{Date of current version 8 April. 2021.}
	\thanks{(Corresponding author: Wei Wang.)}
}

%
%

\markboth{Accepted by IEEE TRANSACTIONS ON PATTERN ANALYSIS AND MACHINE INTELLIGENCE on 31 March~2021}%
{Shell \MakeLowercase{\textit{et al.}}: Bare Advanced Demo of IEEEtran.cls for IEEE Computer Society Journals}
%



\IEEEtitleabstractindextext{%
\begin{abstract}
	
	Eigendecomposition of symmetric matrices is at the heart of many computer vision algorithms. However, the derivatives of the eigenvectors tend to be numerically unstable, whether using the SVD to compute them analytically or using the Power Iteration (PI) method to approximate them. This instability arises in the presence of eigenvalues that are close to each other. This makes integrating eigendecomposition into deep networks difficult and often results in poor convergence, particularly when dealing with large matrices.
	
	While this can be mitigated by partitioning the data into small arbitrary groups, doing so has no theoretical basis and makes it impossible to exploit the full power of eigendecomposition. In previous work, we mitigated this using SVD during the forward pass and PI to compute the gradients during the backward pass. However, the iterative deflation procedure required to compute multiple eigenvectors using PI tends to accumulate errors and yield inaccurate gradients. Here, we show that the Taylor expansion of the SVD gradient is theoretically equivalent to the gradient obtained using PI without relying in practice on an iterative process and thus yields more accurate gradients. We demonstrate the benefits of this increased accuracy for image classification and style transfer.
	
\end{abstract}
\begin{IEEEkeywords}
Eigendecomposition, Differentiable SVD, Power Iteration, Taylor Expansion.
\end{IEEEkeywords}}

\maketitle

\IEEEdisplaynontitleabstractindextext

%
\IEEEpeerreviewmaketitle

\section{Introduction}

In this paper, we focus on the eigendecomposition of symmetric matrices, such as covariance matrices, in a robust and differentiable manner. The eigenvectors and eigenvalues of such matrices are widely used in computer vision to perform tasks such as image classification~\cite{Lei18,Li17a,Li18i,Huang19,Yu17a}, image segmentation~\cite{Carreira12,Ionescu15,Pan19a}, generative  networks~\cite{Chiu19,Siarohin18,Miyato18,Li17d,Cho19}, graph matching~\cite{Zanfir18c,Dang18a,Yi18a}, object pose estimation\cite{Lepetit09,Ferraz14} and style transfer~\cite{Cho19,Chiu19}. 

In practice, eigendecomposition is often performed by Singular Value Decomposition (SVD) because it is more stable than other approaches. Although robustly calculating the derivatives of the resulting eigenvalues is relatively straightforward~\cite{Lewis96}, computing those of the eigenvectors in a numerically stable manner remains an open problem. This is because, even though the eigenvectors are analytically differentiable with respect to the matrix coefficients, their partial derivatives can become uncontrollably large when two eigenvalues are close to each other: These derivatives depend on a matrix $\widetilde{\mK}$ with elements
\begin{equation}
	\begin{aligned}
		\widetilde{\mK}_{i j}=\left\{\begin{array}{ll}{\nicefrac{1}{(\lambda_{i}-\lambda_{j})},} & {i \neq j} \\ {0,} & {i=j}\end{array}\right. \; ,
	\end{aligned}
	\label{eq: mb_k}
\end{equation}
where $\lambda_i$ denotes the $i^{\rm th}$ eigenvalue of matrix being decomposed~\cite{Ionescu15}. Thus, when two eigenvalues are very close, the derivatives become very large and can cause overflows. This makes integrating SVD into deep networks prone to numerical instabilities and may result in poor convergence of the training process. 
When only the eigenvector associated to the largest eigenvalue is needed, this instability can be addressed using the Power Iteration (PI) method~\cite{Nakatsukasa13}. In its standard form, PI relies on an iterative procedure to approximate the dominant eigenvector of a matrix starting from an initial estimate of this vector. This has been successfully used for graph matching~\cite{Zanfir18c} and spectral normalization in generative models~\cite{Miyato18}. 

\begin{figure*}[t!]
	\centering
	\begin{tabular}{cc}
		\includegraphics[width=0.45\linewidth]{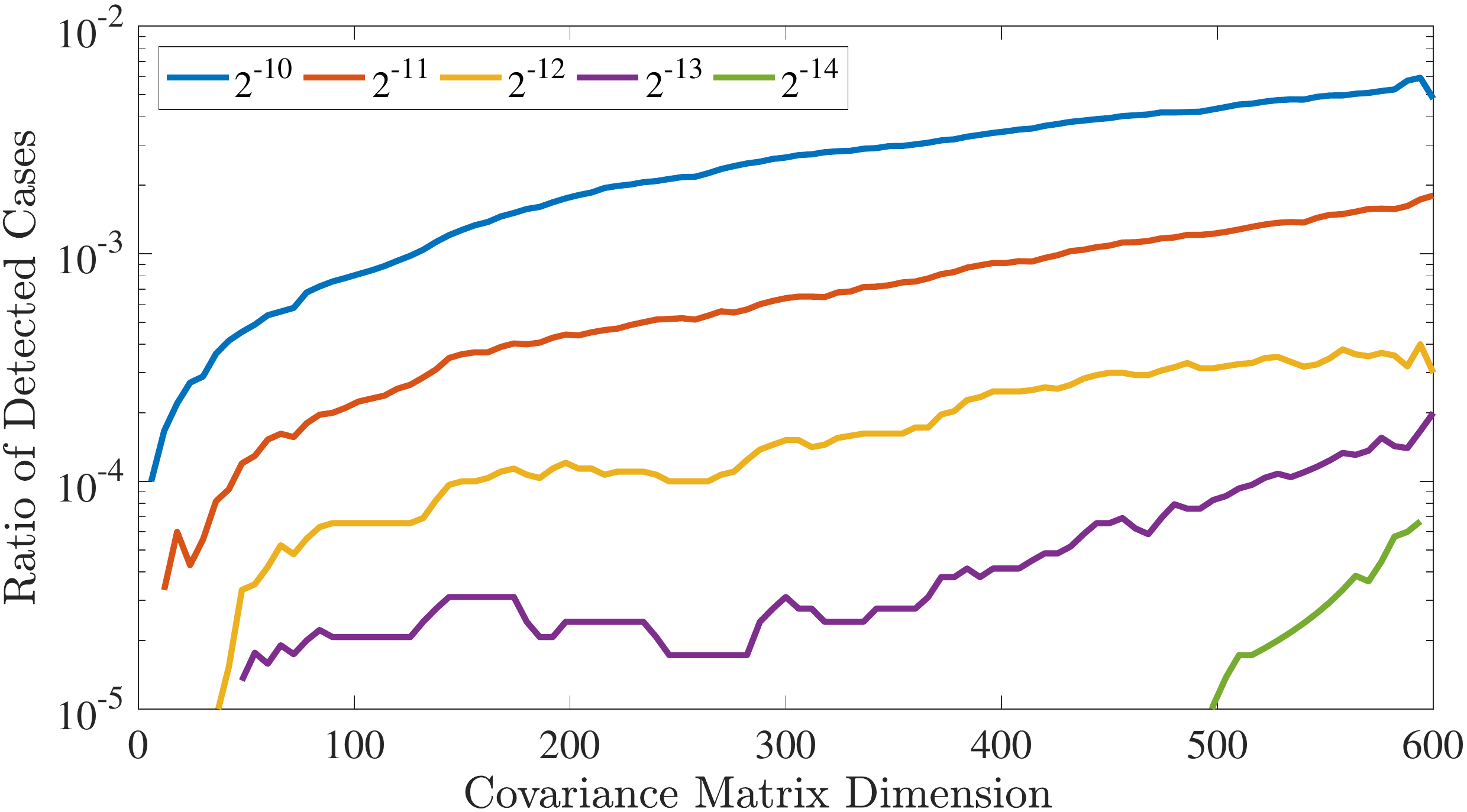}&
		\includegraphics[width=0.45\linewidth]{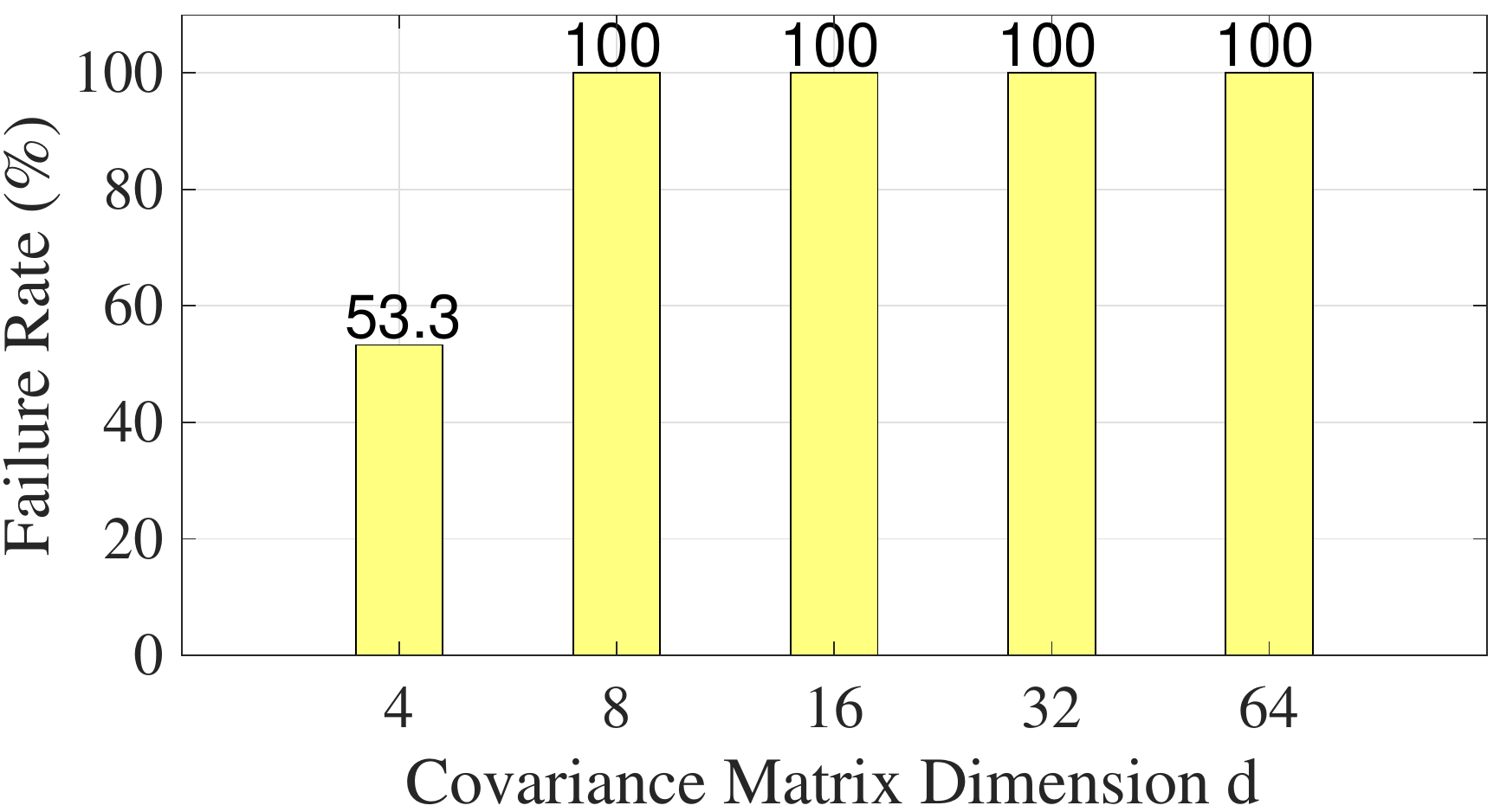}\\
		(a)&(b)
	\end{tabular}
	\vspace{-2mm}
	\caption{\small {\bf Influence of the covariance matrix size.} {\bf (a)} Probability that the difference between two eigenvalues of a covariance matrix is smaller than a threshold---$2^{-10}, \cdots, 2^{-14}$---as a function of its dimension. To compute it for each dimension between 6 and 600, we randomly generated 10,000 covariance matrices and counted the proportion for which at least two eigenvalues were less than a specific threshold from each other. Given the dimension $d$, we randomly sampled $n{=}2d$ data points, $\mX^{d{\times}n}$, whose row-wise mean is 0. The covariance matrix is then obtained by computing $\mX \mX^{\top}$. {\bf (b)} Rate at which training fails for ZCA normalization on CIFAR10. For each dimension, we made 18 attempts.
	}
	\label{fig:ratio-close-eigen}
	\vspace{-2mm}
\end{figure*}

In theory, when {\it all} the eigenvectors are needed, PI can be used in conjunction with a {\it deflation} procedure~\cite{Burden81}. This involves computing the dominant eigenvector, removing the projection of the input matrix on this vector, and iterating. Unfortunately, in practice, this procedure is subject to large round-off errors that accumulate and yields inaccurate eigenvectors and gradients. Furthermore, the results are sensitive to the number of iterations and to how the vector is initialized at the start of each deflation step. Finally, convergence slows down significantly when the ratio between the dominant eigenvalue and the others becomes close to one.

Fig.~\ref{fig:ratio-close-eigen}(a) illustrates this problem and shows that it becomes acute when dealing with large matrices: As the dimension of the matrix grows, so does the probability that at least two eigenvalues will be close to each other.  For example, with a dimension of only 150, there is more that a 0.1\% chance that two eigenvalues will be within $2^{-10}$ of each other, so that some of the $\widetilde{\mK}_{i j}$ are larger than $2^{10}$ according to Eq.~\ref{eq: mb_k} and can trigger numerical instability. As these matrices are generated for all mini-batches during the training procedure of a typical deep network, this adds up to a very high probability of running into this problem during a typical training run. For example, the experiment depicted by Fig.~\ref{fig:ratio-close-eigen}(b) involves covariance matrices of dimension ranging from 4 to 64 to perform ZCA whitening~\cite{Bell97} on the CIFAR10 dataset. When the dimension is 4, the training fails more than 50\% of the time and, when it is larger, all the time. We will discuss this in more detail in the results section.

In practice, heuristics are often used to overcome this problem. For instance, approximate eigenvectors can be learned and a regularizer used to force them to be orthogonal~\cite{Cho19}. The resulting vectors, however, may not be real eigenvectors and we have observed that this yields suboptimal performance. More importantly, while approximate eigenvectors might be acceptable for style transfer as in~\cite{Cho19}, other applications, such as decorrelated batch normalization for image classification~\cite{Huang19,Lei18}, require acccurate eigenvectors. Gradient clipping is another heuristic that can be employed to prevent gradient explosion, but it may affect training and yield inferior performance. A popular way around these difficulties is to use smaller matrices, for example by splitting the feature channels into smaller groups before computing covariance matrices~\cite{Lei18} for ZCA whitening~\cite{Kessy18,Bell97}. This, however, imposes arbitrary limitations on learning, which also degrade performance.

In an earlier conference paper~\cite{Wang19b}, we tackled this by relying on PI during backpropagation while still using regular SVD~\cite{Nakatsukasa13} during the forward pass and for initialization purposes. This allowed us to accurately compute the eigenvectors and remove instabilities for eigenvectors associated to large eigenvalues. This, however, does not address the PI error accumulation problem discussed above.

In this paper, we introduce a novel approach to computing the eigenvector gradients. It relies on the Taylor expansion~\cite{Roy90} of the analytical SVD gradients~\cite{Ionescu15}. Our key insight is that, in theory, the gradients of the SVD computed using PI also are the Taylor expansion of the SVD gradients. However, Taylor expansion does {\it not} require an iterative process over the eigenvectors and is therefore not subject to round-off errors. Ultimately, we therefore use SVD during the forward pass as in our earlier work~\cite{Wang19b}, but we replace PI by Taylor expansion when computing the gradients during backpropagation. We will use the decorrelated batch normalization task~\cite{Huang19,Wang19b} to show that this not only yields more accurate gradients but is also faster because the eigenvectors can be computed in parallel instead of sequentially. The datasets used for this feature decorrelation task are CIFAR10/100, whose image size is 32$\times$32, and Tiny ImageNet, whose image size is 64$\times$64. Both our previous work and our new approach complete training without gradient explosion and converge. The code is available at \url{https://github.com/WeiWangTrento/Robust-Differentiable-SVD}.

To demonstrate the applicability of our approach, we will show that it translates to better image classification performance than in~\cite{Wang19b} and than using a gradient-clipping heuristic in a decorrelated batch normalization layer~\cite{Lei18}. It also delivers improved style transfer when incorporated in the pipelines of~\cite{Chiu19,Cho19}. Furthermore, to evidence scalability, we will use the ImageNet dataset~\cite{Russakovsky15}, whose image size is $256{\times}256$, to test our new approach to perform second-order pooling~\cite{Ionescu15,Li17a}. Not only does our approach eliminate the gradient explosion problem, it also converges for those larger images whereas our earlier approach does not. We chose these three tasks because they all require an SVD of the covariance matrices.

Our contribution is therefore an approach to computing eigenvector gradients for covariance matrices that is fast, accurate, stable, and easy to incorporate in a deep learning framework.

\section{Related Work}

\begin{figure*}[t]
	\centering
	\includegraphics[width=\linewidth]{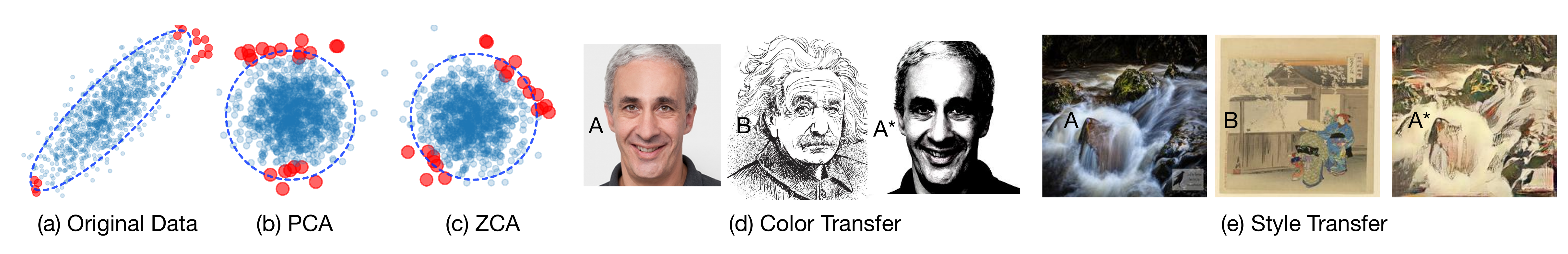}
	\vspace{-6mm}
	\caption{{\bf Applications of differentiable eigendecomposition: ZCA normalization \& color/style transfer} 
		(a) Original $n$ data points $\mX\in\R^{2{\times}n}$, whose two dimensions are highly correlated. 
		(b) PCA whitening removes the correlations, but simultaneously rotates the data, as indicated by the red points. For PCA, the whitening matrix is $\mS_{PCA}=\Lambda^{-\nicefrac{1}{2}}\mV^T$, where $\Lambda$ and $\mV$ denote the diagonal eigenvalue matrix and eigenvector matrix of the covariance matrix $\mX\mX^{\top}$. The points are transformed as $\mX'=\mS_{PCA}\mX$.
		(c) By contrast, ZCA whitening, which also decorrelates the data, preserves the original data orientation. For ZCA, the whitening matrix is $\mS_{ZCA}=\mV\Lambda^{-\nicefrac{1}{2}}\mV^T$, and the points are also transformed as $\mX'=\mS_{ZCA}\mX$.
		Because of its unique property, ZCA whitening can used in a decorrelated batch normalization layer to decorrelate the features~\cite{Huang19}.
		(d) Color transfer can be achieved by first whitening the pixel values in image $A$ by multiplying them with $\mV_A\Lambda_A^{-\nicefrac{1}{2}}\mV_A^T$. The whitened pixels are then colored according to image B by multiplying them with $\mV_B\Lambda_B^{\nicefrac{1}{2}}\mV_B^T$. 
		(e) Style transfer can be achieved by doing similar operations to those performed in color transfer. The difference is that the whitening and coloring transformations are performed at the level of deep feature maps instead of raw pixels. See Section~\ref{sec: style-transfer} for more detail.
	}
	\label{fig: zca-pca}
	\vspace{-4mm}
\end{figure*}

In traditional computer vision, SVD has typically been used within algorithms that do not require its gradient, for instance to obtain an algebraic solution to a least-square problem, as in 6D pose estimation~\cite{Lepetit09,Ferraz14}. 

\subsection{Incorporating SVD in a Deep Network}

In the deep learning context, however, training a network that relies on performing an SVD in one of its layers requires differentiating this process. While the gradient of SVD has an analytical form~\cite{Ionescu15,Papadopoulo00a}, it is numerically unstable and tends to infinity ({\ie} gradient explosion) when two singular values are close to each other, as will be further discussed in Section~\ref{sec:method}. The simplest approach to tackling this is to work in double precision~\cite{Li17a}, which copes better with small singular value differences than single precision that quickly round off this difference to 0. For instance, $1.40{\times}10^{-45}$ can be represented correctly in double precision ({\ie} float64) while it will be rounded to 0 in single precision ({\ie} float32). This, however, solves neither the problem of gradient explosion, as dealing with gradients larger than $3.41 {\times} 10^{38}$ is impractical, nor that of truly equal singular values. 

As a consequence, other mitigation measures have been proposed. For example, the risk of gradient explosion can be reduced by working with smaller submatrices~\cite{Lei18} to bypass two equal singular values. This, however, involves an artificial modification of the original problem, which may result in inferior performance. Another approach therefore consists of bypassing the SVD altogether. This was achieved in our previous work~\cite{Dang18a} by designing a loss term that reflects the SVD equation. However, this is only applicable when one needs to exploit a single singular vector. In~\cite{Wang19b}, we explored the use of PI to compute gradients, but it suffers from accumulated round-off errors in the deflation loop, which leads to inaccurate gradients. Another approach which may mitigate this problem is subspace iteration~\cite{Rutishauser70}, which can be viewed as a generalization of PI to compute a subset of the eigenvalues simultaneously. However, this approach still requires deciding the dimension of the subset, and thus, as with PI, a deflation loop is needed to compute all eigenvalues. Furthermore, subspace iteration relies on QR decomposition~\cite{Andrew14,Liang19a}, which itself involves an iterative process, making the overall procedure time consuming. 

When dealing with covariance matrices, other workarounds have been proposed. For example, for second-order pooling~\cite{Lin15}, which typically relies on computing the product $\mV \Lambda^{-\nicefrac{1}{2}}\mV^{\top}$, with $\mV$ the matrix of eigenvectors of a covariance matrix and $\Lambda$ that of eigenvalues, several methods~\cite{Lin17a,Li18i,Huang19} exploit the Newton-Schulz technique to directly approximate the matrix product, without having to explicitly perform eigendecomposition. Interestingly, this approximation was shown to sometimes outperform the accurate value obtained via SVD for classification purposes~\cite{Li18i}. However, this cannot be generalized to all the tasks involving SVD of covariance matrices, especially ones that require accurate eigenvectors and eigenvalues.

\subsection{Deep Learning Applications of SVD}

Principal component analysis (PCA) whitening~\cite{Bell97} yields decorrelated features. Similarly zero-phase component analysis (ZCA) whitening~\cite{Kessy18} also yields  a decorrelation while retaining the original orientations of the data points, as illustrated in Figure~\ref{fig: zca-pca} (b) and (c). In particular, ZCA whitening~\cite{Kessy18} has been used to design a decorrelated batch normalization layer~\cite{Lei18}, acting as an alternative to Batch Normalization~\cite{Ioffe15}. 
It involves linearly transforming the feature vectors so that their covariance matrix becomes the identity matrix and that they can be considered as decoupled. One difficulty however is that all eigenvectors and eigenvalues are required, which makes a good test case for our approach.

ZCA whitening has also been used for style transfer~\cite{Cho19}. It has been shown that an image's style can be encoded by the covariance matrix of its deep features~\cite{Li17c,Gatys16a,Gatys15}, and that transferring that style from the source image to the target image could be achieved with the help of SVD. Style transfer then involves two steps, whitening and coloring. In the whitening step, ZCA whitening is applied to the features of the source image $A$ to remove its style, using the transformation matrix $\mV_A\Lambda_A^{-\frac{1}{2}}\mV_A^{\top}$.
The resulting whitened image is then colored to the style of a target image $B$ via the transformation matrix $\mV_B\Lambda_B^{\frac{1}{2}}\mV_B^{\top}$~\cite{Chiu19}.
Note that the sign of the exponent of the eigenvalue matrix $\Lambda$ is opposite for the whitening and coloring transformation matrices. 
As shown in Figure~\ref{fig: zca-pca} (d), applying whitening and coloring transformations to the raw pixels transfers the colors from image $B$ to  image $A$. By contrast, applying these transformations to the deep features transfers the style from image $B$ to image $A$.

In~\cite{Cho19}, the ZCA process was circumvented by relying on a deep network to learn the ZCA whitening effect via regularizers. This, however, may negatively impact the results, because it only approximates the true transformations. In~\cite{Wang19b}, we proposed instead to use PI to compute the SVD gradient during backpropagation, while still relying on standard SVD in the forward pass. While this addresses the numerical instabilities, it comes at the cost of reducing the accuracy of the eigenvector gradient estimates because of the round-off errors resulting from the iterative deflation process. Here, we show that this iterative process can be avoided by exploiting the Taylor expansion of the SVD gradient.

Moreover, SVD can also be used for second-order pooling~\cite{Ionescu15}. This operation performs a covariance-based pooling of the deep features instead of using a traditional pooling strategy, such as max or average pooling. Similarly to ZCA whitening, second-order pooling relies on computing a covariance matrix acting as global image representation. This matrix is then typically transformed as  $\mV\Lambda^{\alpha}\mV^{\top}$, where $\alpha$ is usually set to $\frac{1}{2}$~\cite{Li17a}. Second-order pooling is typically implemented after the last convolutional layer of the deep network~\cite{Li17a, Li18i}.

\section{Taylor Expansion for SVD Gradients}
\label{sec:method}

Note that our Taylor expansion method not only outperforms our previous PI based method~\cite{Wang19b}, but also handles the case where two eigenvalues are truly equal.

\begin{table}[!htb]
	\caption{Notation.}
	\label{tab: notation}
	\vspace{-3mm}
	\centering
	\begin{tabular}{c|l}
		\hline
		$\mX{\in}\R^{d \times n}$	& Matrix of $n$ feature points in dimension $d$. \\ 
		$\mu{=}\bar{\mX}$ & Row-wise mean value of $\mX$. \\ 
		$\mI$ & Identity matrix. \\ 
		$\bf{1}{\in}\R^{d \times 1}$ & Vector with all elements equal to 1. \\ 
		$\mM{\in}\R^{d \times d}$	& Covariance matrix of $\mX$. \\ 
		$\mM=\mV\Lambda\mV^{\top}$	& SVD of $\mM$; $\Lambda$, $\mV$ denote eigenvalues \& eigenvectors.\\ 
		$\Lambda_{diag} {=} diag(\Lambda)$& Vector of eigenvalues. \\ 
		$\epsilon$	& Small positive value. \\
		$\lambda_i$	& $i$-th eigenvalue, ranked in descending order.\\
		$\vv_i$	& $i$-th eigenvector associated with $\lambda_i$.\\
		$\mA\circ\mB$	& Element-wise product between $A$ and $B$. \\ 
		\hline
	\end{tabular}
\end{table}

We now introduce our approach to using the Taylor expansion of the SVD gradient to compute the gradient of eigenvectors estimates with respect to the matrix coefficients. To this end, we first review the analytical form of the gradient, derive its Taylor expansion, and then show the close relationship between this expansion and traditional  PI. We conclude the section by highlighting some of the theoretical benefits of our approach over PI and gradient-clipping. We will demonstrate the practical ones in the results section.

Table~\ref{tab: notation} summarizes the notation we will use from now on. Below, we assume that the eigenvalues are ranked in the descending order with $\lambda_{1}{\geq}\lambda_{2}{\geq}{\dots}{\geq}\lambda_{n}$.

\subsection{SVD Gradient for Covariance Matrices}
\label{sec: svd-cov}
The SVD of a covariance matrix $\mM {\in} \R^{d {\times} d}$ can be expressed as $\mM{=}\mV\Lambda\mV^{\top}$, where $\mV$ is the matrix of eigenvectors such that $\mV\mV^{\top}{=}\mI$, and $\Lambda$ is the diagonal matrix containing the eigenvalues. Our goal is to perform end-to-end training of a deep network that incorporates eigendecomposition as one of its layers. To this end, we need to compute the gradient of the loss function $\mL$, depending on the SVD of $\mM$, w.r.t. the matrix $\mM$ itself.
As shown in~\cite{Ionescu15}, the partial derivatives can be computed as
\begin{equation}
	\frac{\partial \mL}{\partial \mM} {=} \mV\left( \left( \widetilde{\mK}^{\top} {\circ} \left(\mV^{\top}\frac{\partial \mL}{\partial \mV} \right) \right){+}\left( \frac{\partial \mL}{\partial \Lambda}_{diag} \right) \right) \mV^{\top}\;,
	\label{eq: svd-cov}
\end{equation}
with $\widetilde{\mK}$ a matrix whose ($i,j$)-th element is given by 
\begin{equation}
	\widetilde{\mK}_{i,j} = 
	\begin{cases}
		\frac{1}{(\lambda_i - \lambda_j)},& i \neq j \\
		0,              & i=j
	\end{cases}\;.
	\label{eq: unstable_term_eig}
\end{equation}

\noindent Introducing Eq.~\ref{eq: unstable_term_eig} into Eq.~\ref{eq: svd-cov} yields
\begin{equation}
	\frac{\partial \mL}{\partial \mM} 
	{=}\sum_{i=1}^n\sum_{j \neq i}^{n}\frac{1}{\lambda_{i} {-}\lambda_{j}}\vv_{j}\vv_{j}^{\top}\frac{\partial \mL}{\partial \vv_{i}}\vv_{i}^{\top}
	{+} \sum_{i=1}^n\frac{\partial \mL}{\partial \lambda_i}\vv_i\vv_i^{\top}\;.
	\label{eq: svd-org-form}
\end{equation}
The instability of this analytical gradient arises from the term $\lambda_i{-}\lambda_j$ in the denominator. If two eigenvalues are equal, or almost equal, the resulting gradient will tend to infinity and explode. To solve this problem, we rely on its Taylor expansion.

\subsection{Taylor Expansion of the Gradient}
Let us now consider the Taylor expansion of $\widetilde{\mK}_{i,j}$ in Eq.~\ref{eq: unstable_term_eig}. To this end, recall that the $K$-th degree Taylor expansion of $f(x)=\frac{1}{1-x}, \ \ x {\in} [0, 1)$, at a point $x_0{=}0$ is  given by 
\begin{align}
	\label{equation:eq: expansion}
	&f(x)=1+x+x^{2}+\cdots +x^{K}+ R_{K+1}(x),\\
	\label{eq: reminder}
	&R_{K+1}(x) = \frac{x^{K+1}}{1-x}\;,
\end{align}
where $R_{K+1}(x)$ is the remainder of the expansion. Then, if we substitute $x$ in this expression with $\lambda_j/\lambda_i$, we can write
\begin{equation}
	\frac{1}{1{-}\left(\nicefrac{\lambda_j}{\lambda_i}\right)} 
	\approx 1{+}\left(\frac{\lambda_j}{\lambda_i}\right){+}\left(\frac{\lambda_j}{\lambda_i}\right)^{2} {+} \cdots {+}\left(\frac{\lambda_j}{\lambda_i}\right)^{K}\;.
	\label{eq: expansion-svd-term2}
\end{equation}
We now observe that
\begin{equation}
	\widetilde{\mK}_{i,j} =  \frac{1}{\left(\lambda_i {-} \lambda_j \right)}  = \frac{1}{\lambda_i}\cdot\frac{1}{1-\left(\nicefrac{\lambda_j}{\lambda_i}\right)}\;.
	\label{eq:kt_2}
\end{equation}
Therefore, exploiting Eq.~\ref{eq:kt_2} together with Eq.~\ref{eq: expansion-svd-term2}, we can write the Taylor expansion of $\widetilde{\mK}_{i,j}$ and its remainder as
\begin{equation}
	\widetilde{\mK}_{i,j}
	= \frac{1}{\lambda_i}\left(1{+}\left(\frac{\lambda_j}{\lambda_i}\right){+}\cdots{+}\left(\frac{\lambda_j}{\lambda_i}\right)^{K} \right) 
	{+}R_{K+1} \;,
	\label{eq: expansion-svd-sym}
\end{equation}
\begin{equation}
	R_{K+1} {=} \frac{1}{\left(\lambda_i {-} \lambda_j \right)} \left(\frac{\lambda_j}{\lambda_i}\right)^{K+1} \;.
	\label{eq: reminder-ed-talor}
\end{equation}
Let us now inject this into Eq.~\ref{eq: svd-org-form}. To this end, we note that the first term in Eq.~\ref{eq: svd-org-form} can be split into two parts: $j>i$ and $j<i$. We can therefore rewrite it as
\begin{equation}
	\resizebox{0.88\linewidth}{!} 
	{$
		\sum_{i=1}^n \left( \sum_{j > i}^{n}\frac{1}{\lambda_{i}{-}\lambda_{j}}\vv_{j}\vv_{j}^{\top}\frac{\partial \mL}{\partial \vv_{i}}\vv_{i}^{\top} 
		{-} \sum_{j < i}^{n}\frac{1}{\lambda_{j}{-}\lambda_{i}}\vv_{j}\vv_{j}^{\top}\frac{\partial \mL}{\partial \vv_{i}}\vv_{i}^{\top} \right).
		$}
	\label{eq: two-cases}
\end{equation}

\noindent Replacing $\widetilde{\mK}$ in this expression with its Taylor expansion given by Eq.~\ref{eq: expansion-svd-sym} will yield an approximate gradient. To compute it, let us first focus on the approximation of the first part of Eq.~\ref{eq: two-cases} in which $\lambda_i \geq \lambda_j, \ (i {<} j)$. This can be expressed as
\begin{equation}
	\sum_{j > i}^{n} \frac{1}{\lambda_i}\left(1{+}\left(\frac{\lambda_j}{\lambda_i}\right){+}\cdots{+}\left(\frac{\lambda_j}{\lambda_i}\right)^{K} \right) \vv_{j}\vv_{j}^{\top}\frac{\partial \mL}{\partial \vv_{i}}\vv_{i}^{\top}\;.
	\label{eq: svd-taylor-form}
\end{equation}
The approximation for the second part of Eq.~\ref{eq: two-cases}, where $\lambda_{i} {\leq} \lambda_{j}, \ (i {>} j)$, can be obtained in a similar manner.
Note that the difference $\lambda_i {-} \lambda_j$ has disappeared from the gradient, and thus having two equal eigenvalues will not lead to an infinite gradient. Nevertheless, $\lambda_i$ appears in the denominator, and thus the gradient will still be $\infty$ when $\lambda_i{=}0$.

Given that $\mM$ is positive semidefinite, we have $\lambda_{i}{\geq}0$ for any $i$. To prevent $\lambda_i{=}0$, we add a small positive value $\epsilon$ to the diagonal of the matrix $\mM$. We then have $\widetilde{\mM} {=} \mM{+}\epsilon\mI$, which guarantees that $\lambda_{i} {\geq} \epsilon$. Importantly, this $\epsilon$ does not affect the eigenvectors. To show this, recall that the $i$-th eigenvector of a symmetric matrix $\mM$ satisfies
\begin{equation}
	\mM\vv_{i} = \lambda_i \vv_{i}\;,
\end{equation}
with $\lambda_i$ representing the $i$-th eigenvalue. Then, substituting $\mM$ by $\widetilde{\mM}$ yields
\begin{equation}
	\left(\mM {+} \epsilon \mI \right) \vv_{i} {=} \lambda_i \vv_{i} {+} \epsilon \vv_{i} {=} \left( \lambda_i {+} \epsilon \right)\vv_{i} \;.
	\label{eq: epsilon-influence}
\end{equation}
Hence, while the eigenvalue is affected by the value of $\epsilon$, the eigenvector $\vv_{i}$ is not. From this follows the fact that the gradient magnitude, unlike that of Eq.~\ref{eq: svd-cov}, is bounded when approximated using Eq.\ref{eq: svd-taylor-form}, which we now prove. 
To this end, let us consider the approximation of Eq.\ref{eq: svd-taylor-form}. We have
\begin{equation}
	\resizebox{0.85\linewidth}{!} 
	{$
		\begin{aligned}
			& \left\Vert  \sum_{j > i}^{n} \frac{1}{\lambda_i}\left(1{+}\frac{\lambda_j}{\lambda_i}{+}\cdots{+}\left(\frac{\lambda_j}{\lambda_i}\right)^{K} \right) \vv_{j}\vv_{j}^{\top}\frac{\partial \mL}{\partial \vv_{i}}\vv_{i}^{\top} \right\Vert
			\\
			{\leq}  &\sum_{j > i}^{n}  \frac{1}{\lambda_i} \left(1{+}\frac{\lambda_j}{\lambda_i}
			{+}\cdots{+}\left(\frac{\lambda_j}{\lambda_i}\right)^{K} \right)
			\left\Vert\vv_{j}\vv_{j}^{\top} \right\Vert \left\Vert \frac{\partial \mL}{\partial \vv_{i}}\right\Vert \left\Vert\vv_{i}^{\top} \right\Vert \; ,
			\\ 
			{\leq} & \sum_{j > i}^{n} \frac{K{+}1}{\lambda_i} 
			\left\Vert \frac{\partial \mL}{\partial \vv_{i}} \right\Vert 
			{\leq} \frac{n(K{+}1)}{\epsilon} \left\Vert \frac{\partial \mL}{\partial \vv_{i}} \right\Vert \; .
		\end{aligned}
		$}
	\label{eq: svd-taylor-form-bound}
\end{equation}
Note that the resulting bound  $\frac{n(K+1)}{\epsilon}$ is a constant that is independent from $\lambda_i$ and $\lambda_j$. It only depends on the value of the 2 hyperparameters, $K$ and $\epsilon$. A large $\epsilon$ and a small $K$ will yield a low bound and will constrain the gradient magnitude more strongly. However, more energy of the gradient will be trimmed off. In Section~\ref{sec: bound} on the appendix, we will use a concrete numerical example to show this has virtually no impact on the gradient direction. 

\subsection{Relationship with Power Iteration Gradients}
\label{sec: pi}

The Taylor expansion-based gradient derived above and the PI-based one studied in our previous work~\cite{Wang19b} are related as follows.

\noindent {\bf Proposition. }{\it The $K$-th degree Taylor expansion of the gradient is equivalent to computing the gradient of $K{+}1$ power iterations when PI is not involved in the forward pass.} To prove this, we first review the PI gradient~\cite{Wang19b}.

Let $\mM$ be a covariance matrix. To compute its leading eigenvector $\vv$, PI relies on the iterative update
\begin{equation}
	\vv^{(k)} = \frac{\mM\vv^{(k-1)}}{\| \mM\vv^{(k-1)} \|}\;,
\end{equation}
where $\| {\cdot} \|$ denotes the $\ell_2$ norm.
The PI gradient can then be computed as~\cite{Mang17}
\begin{equation}
	\resizebox{0.68\linewidth}{!} 
	{$
		\begin{aligned}
			&\frac{\partial \mL}{\partial \mM} {=}\sum_{k=0}^{K-1} \frac{\left(\mI{-}\vv^{(k+1)} \vv^{(k+1)\top}\right)}{\left\|\mM \vv^{(k)}\right\|} \frac{\partial \mL}{\partial \vv^{(k+1)}} \vv^{(k)\top}, \\
			&\frac{\partial \mL}{\partial \vv^{(k)}} {=}\mM \frac{\left(\mI {-}\vv^{(k+1)} \vv^{(k+1)\top}\right)}{\left\|\mM \vv^{(k)}\right\|} \frac{\partial \mL}{\partial \vv^{(k+1)}}.
		\end{aligned}
		$}
	\label{eq:pi_gradient}
\end{equation}

\noindent In the foward pass, the eigenvector $\vv$ can be computed via SVD. Feeding it as initial value in PI will yield 
\begin{equation}
	\vv =\vv^{(0)} \approx\vv^{(1)} \approx \cdots \approx\vv^{(k)} \cdots \approx\vv^{(K{+}1)}
	\label{eq: premise}
\end{equation}
Exploiting this in Eq.~\ref{eq:pi_gradient} and introducing the explicit form of $\frac{\partial \mL}{\partial \vv^{(k)}}, \; k=1,2, \cdots, K{+}1$, into $\frac{\partial \mL}{\partial \mM}$ lets us write
\begin{equation}
	\resizebox{0.88\linewidth}{!} 
	{$
		\frac{\partial \mL}{\partial \mM}
		{=} \left( \frac{\left(\mI{-}\vv \vv^{\top}\right)}{\left\|\mM \vv\right\|}  {+}
		\frac{\mM \left(\mI{-}\vv \vv^{\top}\right)}{\left\|\mM \vv\right\|^{2}}  {+}
		\cdots
		\frac{\mM^{K} \left(\mI{-}\vv \vv^{\top}\right)}{\left\|\mM \vv\right\|^{K+1}} \right) \frac{\partial \mL}{\partial \vv}
		\vv^{\top}.
		$}
	\label{eq:pi_pytorch}
\end{equation}	

\noindent Eq.~\ref{eq:pi_pytorch} is the form adopted in~\cite{Zanfir18c} to compute the ED gradients. Note that we have  
\begin{align}
	\label{eq: term_k}
	&\mM^{k}  {=} \mV \Sigma^{k} \mV^{\top}  {=}  \lambda_{1}^{k}\vv_{1}\vv_{1}^{\top} {+} \lambda_{2}^{k}\vv_{2}\vv_{2}^{\top} {+} \cdots {+} \lambda_{n}^{k}\vv_{n}\vv_{n}^{\top},
	\\
	\label{eq: lambda_k}
	&\left\|\mM\vv\right\|^k = \left\| \lambda \vv \right\|^k= \lambda^k\;.
\end{align}

\noindent Let $\vv_1$ and $\lambda_1$ be the dominant eigenvector and eigenvalue of $\mM$.
Introducing Eq.~\ref{eq: term_k} \& \ref{eq: lambda_k} into Eq.~\ref{eq:pi_pytorch} let us re-write the gradient as
\begin{equation}
	\resizebox{0.88\linewidth}{!} 
	{$
		\begin{aligned} 
			\frac{\partial \mL}{\partial \mM}
			&{=}\left( \frac{\sum_{i=2}^{n}\vv_{i}\vv_{i}^{\top}}{\lambda_1} {+}
			\frac{\sum_{i=2}^{n}\lambda_{i}\vv_{i}\vv_{i}^{\top}}{ \lambda_{1}^{2}} {+}  \cdots {+}
			\frac{\sum_{i=2}^{n}\lambda_{i}^{K}\vv_{i}\vv_{i}^{\top}}{\lambda_{1}^{K+1}} \right) \frac{\partial \mL}{\partial \vv_1}
			\vv_1^{\top}\\
			&{=}\left(\sum_{i=2}^{n}\left(
			\frac{1}{\lambda_{1}} {+}
			\frac{1}{\lambda_{1}}\left(\frac{\lambda_{i}}{\lambda_{1}}\right)^{1} {+} \cdots {+}
			\frac{1}{\lambda_{1}}\left(\frac{\lambda_{i}}{\lambda_{1}}\right)^{K}
			\right)\vv_{i}\vv_{i}^{\top}
			\right)\frac{\partial \mL}{\partial \vv_{1}}\vv_{1}^{\top}\\
			&{=}\left(\sum_{i=2}^{n} \frac{1}{\lambda_{1}} \left(1 {+}
			\left(\frac{\lambda_{i}}{\lambda_{1}}\right)^{1} {+} \cdots {+}
			\left(\frac{\lambda_{i}}{\lambda_{1}}\right)^{K}
			\right)\vv_{i}\vv_{i}^{\top}
			\right)\frac{\partial \mL}{\partial \vv_{1}}\vv_{1}^{\top}\;.
		\end{aligned}
		$}
	\label{eq: geo-prog-series}
\end{equation}

\noindent Note that Eq.~\ref{eq: geo-prog-series}, derived using PI~\cite{Wang19b}, is similar to Eq.~\ref{eq: svd-taylor-form}, obtained by Taylor expansion. If we set $i{=}1$ in Eq.~\ref{eq: svd-taylor-form}, which represents the derivative w.r.t. the dominant eigenvector $\vv_1$, then the two equations are identical. Thus, for the dominant eigenvector, using the $k^{\rm th}$ degree Taylor expansion is equivalent to performing $k{+}1$ power iterations. In the next section, we will introduce the deflation process which iteratively removes the dominant eigenvector: After removing the dominant eigenvector $\vv_1$ from $\mM$, the second largest eigenvector $\vv_2$ becomes the largest. Then, the statement that using the $k^{\rm th}$ degree Taylor expansion is equivalent to performing $k{+}1$ power iterations remains true for the second largest eigenvector $\vv_2$, and ultimately this equivalence can be applied iteratively for the following eigenvectors.
Note also that our proposition works under the premise that PI is not used in the forward pass. Otherwise, Eq. \ref{eq: premise} will not be satisfied. 

\subsection{PI Gradients vs Taylor Expansion}

\setlength{\textfloatsep}{0pt}
\begin{algorithm}[t!]
	\caption{ZCA whitening with SVD-PI}
	\label{alg: svd-pi}
	{Centralize $\mX$: $\mu{\leftarrow}\bar{\mX}$, $\widetilde{\mX}{\leftarrow}\mX{-}\mu\bf{1}^{\top}$}\;
	{Compute Covariance Matrix: $\mM{\leftarrow}\widetilde{\mX}\widetilde{\mX}^{\top}{+}\epsilon I$}\;
	{Initialize running mean and subspace $E_{\mu} {\leftarrow} 0, E_{S} {\leftarrow} \mI$}\;
	{Momentum: $m{\leftarrow}0.1$}\;
	{{\bf Forward pass}: }
	{Standard SVD: $\mV{\Lambda}\mV^{\top}{\leftarrow}\textrm{SVD}(\mM)$; $\Lambda_{diag} {\leftarrow} [\lambda_1,...,\lambda_n], \mV{\leftarrow}\left[\vv_1,...,\vv_n\right],\widetilde{\mM}_0  {\leftarrow} \mM, {rank} {\leftarrow} 1$};
	\For{$i=1:n$}{
		$\vv_i$ $\leftarrow$ Power Iteration $(\widetilde{\mM}_{i-1}, \vv_i)$\;
		$\widetilde{\lambda_i} {\leftarrow} \nicefrac{\vv_i^{\top}\widetilde{\mM}\vv_i}{(\vv_i^{\top}\vv_i)}, 
		\widetilde{\mM}_i {\leftarrow} \widetilde{\mM}_{i-1} {-} \widetilde{\mM}_{i-1}\vv_i\vv_i^{\top}$\;
		{Compute the energy preserved by the top $i$ eigenvalues: $\gamma_i{\leftarrow}\nicefrac{\sum_{k=1}^i\lambda_k}{\sum_{k=1}^n\lambda_k}$}\;
		\eIf{$\lambda_i {\leq} \epsilon$  \  or \  $\nicefrac{\left| \widetilde{\lambda_i}-\lambda_i \right|}{\lambda_i} {\geq} 0.1$  \  or  \  $\gamma_i {\geq} (1{-}0.0001)$}{
			break\;
		}{${rank} {\leftarrow} i$, $\widetilde{\Lambda} {\leftarrow} [\widetilde{\lambda_1}, \cdots, \widetilde{\lambda_i}]$.}
	}
	{\bf Truncate} eigenvectors: $\widetilde{\mV} \leftarrow [\vv_1, ..., \vv_{rank}]$\;
	Compute subspace (whitening transformation matrix): $\mS \leftarrow \widetilde{\mV}(\widetilde{\Lambda})^{-\frac{1}{2}}\widetilde{\mV}^{\top}$\;
	ZCA whitening:
	$\mX \leftarrow \mS\widetilde{\mX}$\;
	Update the running mean and subspace: $E_{\mu} {\leftarrow} m {\cdot} \mu {+} (1{-}m) {\cdot} E_{\mu}$, $E_{\mS} {\leftarrow} m {\cdot} \mS {+} (1{-}m) {\cdot} E_{\mS}$\;
	\KwOut{affine transformation: $\mX{\leftarrow} \gamma \mX{+}\beta$}
	{{\bf Backward pass}: }
	\For{$i=1:{rank}$}{
		Introduce $\vv_i, \widetilde{\mM}_{i}$ into Eq.~\ref{eq:pi_pytorch} to get the gradient for $\vv_i$\;
	}
\end{algorithm}

We have demonstrated the theoretical equivalence of computing gradients using PI or our Taylor expansion. We now introduce the practical benefits of the latter. To this end, we provide in Alg.~\ref{alg: svd-pi} and~\ref{alg: svd-taylor} the pseudocodes corresponding to ZCA whitening using the Power Iteration method~\cite{Wang19b} and our Taylor expansion approach, respectively. Note that the forward pass of these two algorithms, although seemingly different, mostly perform the same operations. In particular, they both compute a running mean $E_{\mu}$ similar to that of standard batch normalization and a running subspace $E_{\mS}$ based on the eigenvector-based whitening transformation, replacing the running variance of standard batch normalization. The running mean $E_{\mu}$ and running subspace $E_{\mS}$ are then used in the testing phase.

The main difference between the forward pass of the two algorithms arises from the iterative deflation process used by Alg.~\ref{alg: svd-pi}, starting at Line 5. In Alg.~\ref{alg: svd-pi}, the operation
\begin{equation}
	\vv_i {=} \text{Power Iteration}(\widetilde{\mM}_i, \vv_i) 
\end{equation}
has in principle no effect on the forward pass because the initial $\vv_i$ is obtained via SVD. It only serves to save $\widetilde{\mM}_i$ and $\vv_i$ that will be used to compute the gradient during the backward pass. 

After computing the dominant eigenvalue and eigenvector of the matrix using PI, we also need to compute the other eigenvalues. $\widetilde{\mM}$ can be written as $\sum_i^n \lambda_i\vv_{i}\vv_{i}^{\top}$, and we remove the dominant direction from the matrix by computing

\begin{equation}
	\widetilde{\mM} \leftarrow \widetilde{\mM} -  \lambda_1\vv_1\vv_1^{\top} = \widetilde{\mM} -  \widetilde{\mM}\vv_1\vv_1^{\top}\;.
\end{equation}

Then the dominant eigenvalue of the new $\mM$ is the second largest one of the original matrix. We can again compute it using PI. We repeat the operations above and finally obtain all the eigenvalues and eigenvectors. This procedure is referred to as {\it deflation process}.

In practice, removing the dominant eigenvector from matrix $\widetilde{\mM}$ introduces a round-off error. 
Repeating this process then accumulates the round-off errors,
often to the point of resulting in a non-positive semidefinite matrix, which violates the basic assumption made by this approach and degrades the performance. Furthermore, this decreases the accuracy of the resulting $\vv_i$. 

\setlength{\textfloatsep}{0pt}
\begin{algorithm}[t!]
	\caption{ZCA whitening with SVD-Taylor}
	\label{alg: svd-taylor}
	{Centralize $\mX$: $\mu{\leftarrow}\bar{\mX}$, $\widetilde{\mX}{\leftarrow}\mX{-}\mu\bf{1}^{\top}$}\;
	{Compute Covariance Matrix: $\mM{\leftarrow}\widetilde{\mX}\widetilde{\mX}^{\top}{+}\epsilon I$}\;
	{Initialize running mean and covariance matrix $E_{\mu} {\leftarrow} 0, E_{\mM} {\leftarrow} \mI$}\;
	{Momentum: $m{\leftarrow}0.1$}\;
	{{\bf Forward pass}: }
	{Standard SVD: $\mV{\Lambda}\mV^{\top}{\leftarrow}\textrm{SVD}(\mM)$; $\lambda_i{\leftarrow}\max(\lambda_i, \epsilon), i{=}1,2,\cdots, n$ $\Lambda_{diag} {\leftarrow} [\lambda_1, \cdots, \lambda_n],\; \mV{\leftarrow}\left[\vv_1, \cdots, \vv_n\right]$}\;
	Compute subspace (whitening transformation matrix): $\mS \leftarrow \mV (\Lambda)^{-\frac{1}{2}}\mV^{\top}$\;
	ZCA whitening:
	$\mX \leftarrow \mS\widetilde{\mX}$\;
	
	Update the running mean and running covariance matrix: $E_{\mu} {\leftarrow} m {\cdot} \mu {+} (1{-}m) {\cdot} E_{\mu}$, $E_{\mM} {\leftarrow} m {\cdot} \mM {+} (1{-}m) {\cdot} E_{\mM}$\;
	\KwOut{affine transformation: $\mX{\leftarrow} \gamma \mX{+}\beta$}
	{{\bf Backward pass}: }
	{Compute the Taylor expansion of $\widetilde{\mK}$ according to Eq.~\ref{eq: expansion-svd-sym}}\;
	{Compute the gradient using Eq.~\ref{eq: svd-cov}.}
\end{algorithm}

The eigenvalues, which are computed as Rayleigh quotients of the form
\begin{equation}
	\widetilde{\lambda_i} {=} \frac{\vv_i^{\top}\widetilde{\mM_i}\vv_i}{\vv_i^{\top}\vv_i}\;,
	\label{eq: Rayleigh}
\end{equation}
become increasingly inaccurate, diverging from the true eigenvalues $\lambda_i$. Note that these two sources of errors will affect the backward pass, because an increasingly inaccurate $\widetilde{\mM}_i$ will also lead to an increasingly inaccurate gradient, due to their dependency as expressed in Eq.~\ref{eq:pi_pytorch}.

To overcome this, Alg.~\ref{alg: svd-pi} relies on the breaking conditions $\lambda_i {\leq} \epsilon$, to avoid eigenvalues smaller than their theoretical bound $\epsilon$, and $\nicefrac{\left| \widetilde{\lambda_i}-\lambda_i \right|}{\lambda_i} {\geq} 0.1$, to prevent the eigenvalues computed via the Rayleigh quotient of Eq.~\ref{eq: Rayleigh} from differing too much from the SVD ones. Furthermore, the deflation process also terminates if $\gamma_i \geq (1-0.0001)$, that is when the remaining energy in $\widetilde{\mM}$ is less than 0.0001, which improves stability. Altogether, these conditions result in the discarding of small eigenvalues and thus in the gradient of Eq.~\ref{eq: geo-prog-series} being only approximated. By contrast, the Tayor expansion-based Alg.~\ref{alg: svd-taylor} uses the full eigenspectrum of $\mM$, thus yielding a more accurate gradient. Furthermore, it requires no for-loop, which makes it faster than Alg.~\ref{alg: svd-pi}.

We have observed that the eigenvalues computed by standard SVD are not always accurate, even when using double precision, which affects both algorithms. For Alg.~\ref{alg: svd-pi}, this is accounted for by the three breaking conditions in the for-loop. For Alg.~\ref{alg: svd-taylor}, we simply address this by clamping the eigenvalues using $\lambda_i{=}\max(\lambda_i, \epsilon), i{=}1,2,\cdots, n$.

In both algorithms, we use an affine transformation at the end of the forward pass. The affine transformation is a standard operation in a Batch Normalization layer. As explained in~\cite{Ioffe15}, the process of standardization (subtracting the mean and dividing by the standard deviation) throws away important information about the previous layer, which may lead to a decreased ability for the model to learn complicated interactions. This can be resolved by adding two trainable parameters, $\gamma$ and $\beta$, that allow us to re-introduce some information of the previous layer via an affine transformation.

For decorrelated batch normalization, it has also been shown that the stochastic sequences of the mini-batch whitening matrix have a larger diversity than the covariance matrix~\cite{Huang20}. Therefore, instead of computing the mini-batch whitening matrix $\mS$ and its moving average directly, we first compute the mean of the covariance matrices $\mM$, and then $\mS$ given $\mM$ after training, which makes inference more stable. Therefore, in Alg.~\ref{alg: svd-taylor} we compute the running covariance matrices $\mM$ instead of running whitening matrix $\mS$.

\subsection{Practical Limitations of Gradient Clipping}
\label{sec: truncation}
\begin{figure}[t!]
	\centering
	\includegraphics[width=\linewidth]{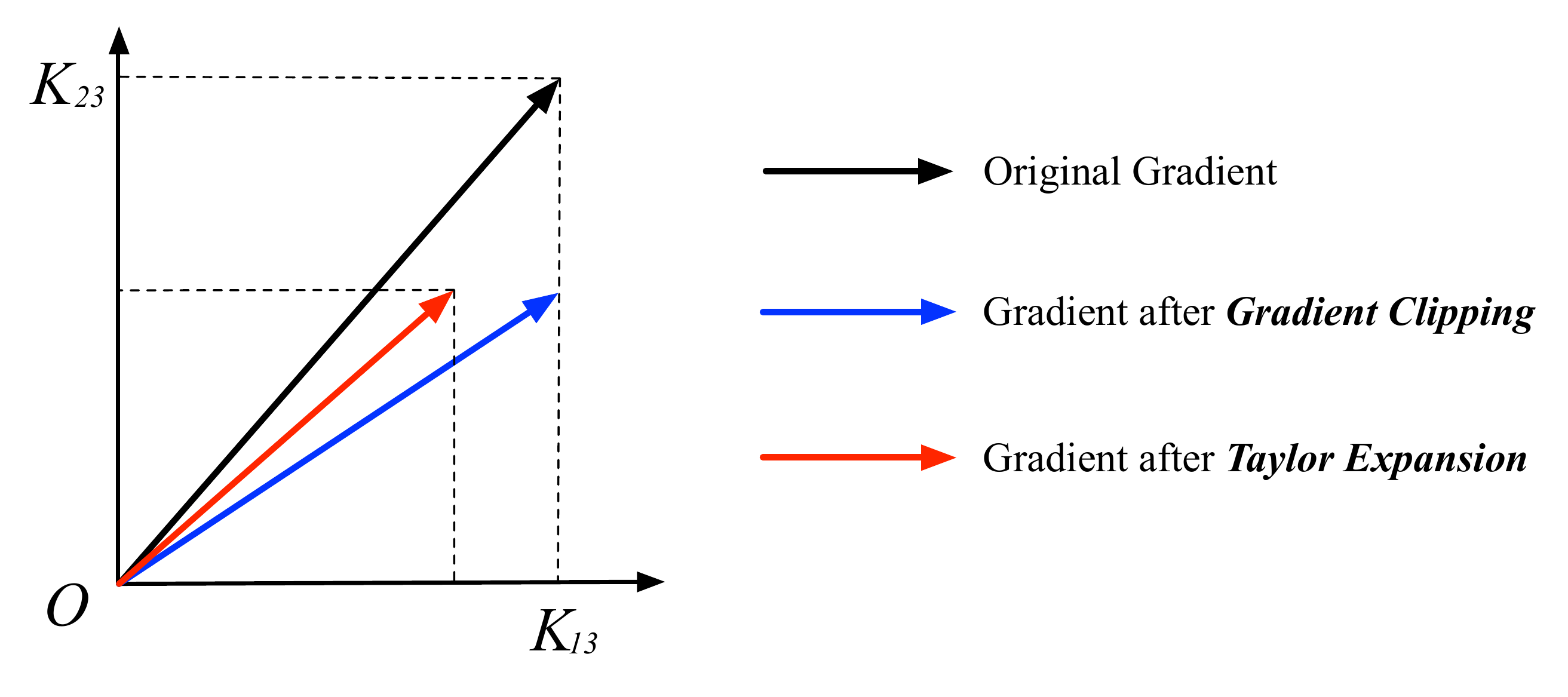}
	\vspace{-8mm}
	\caption{Original gradient descent direction and the ones after gradient clipping and Taylor expansion. We observe that the direction is better preserved with the Taylor expansion.}
	\label{fig: clip}
\end{figure}

\begin{figure*}[t!]
	\centering
	\subfloat[Ratio of $R_{K+1}(x)$ to $f(x)$.]{\includegraphics[width=0.45\linewidth]{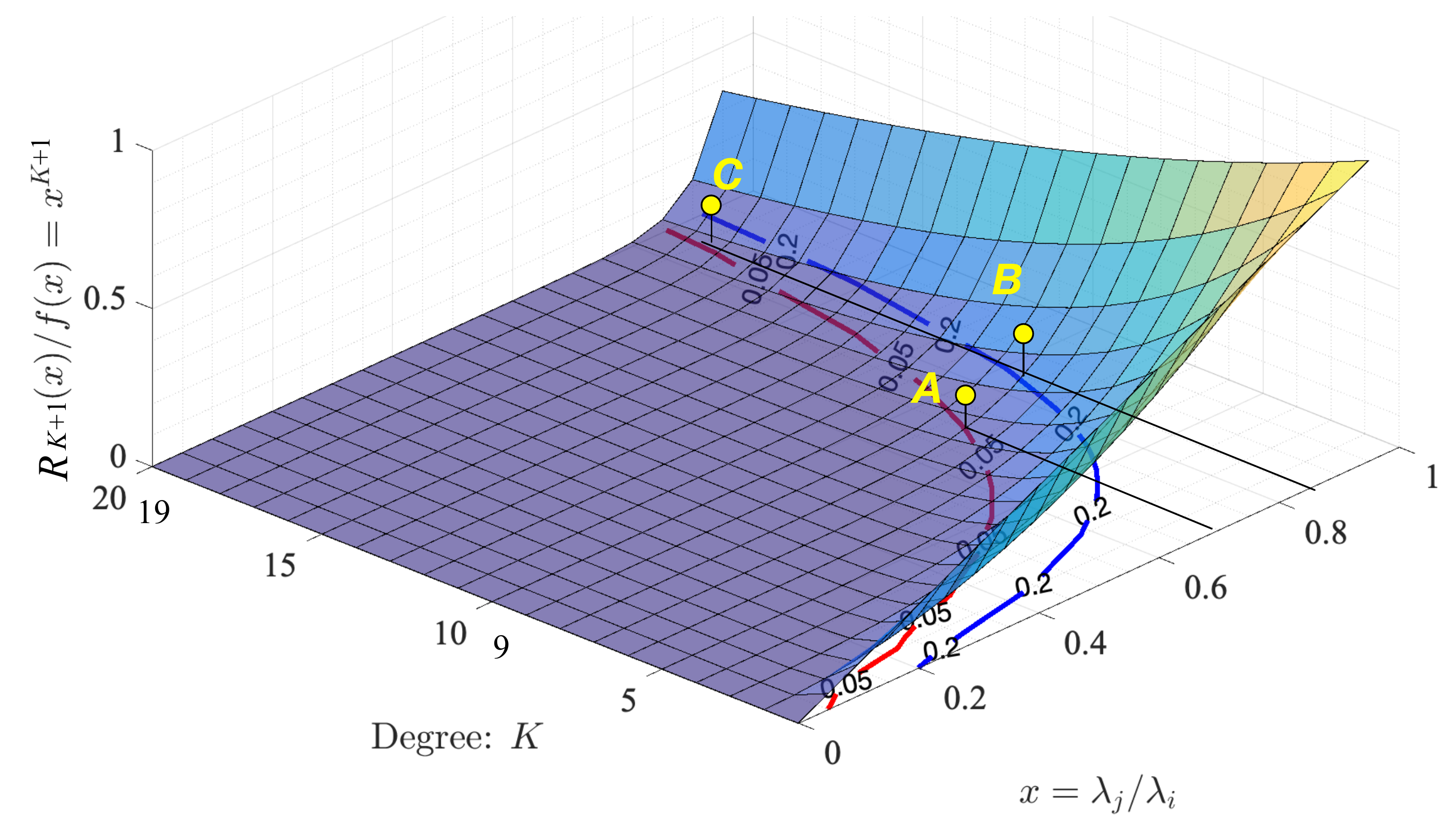} }%
	\quad
	\subfloat[Contour plot]{\includegraphics[width=0.35\linewidth]{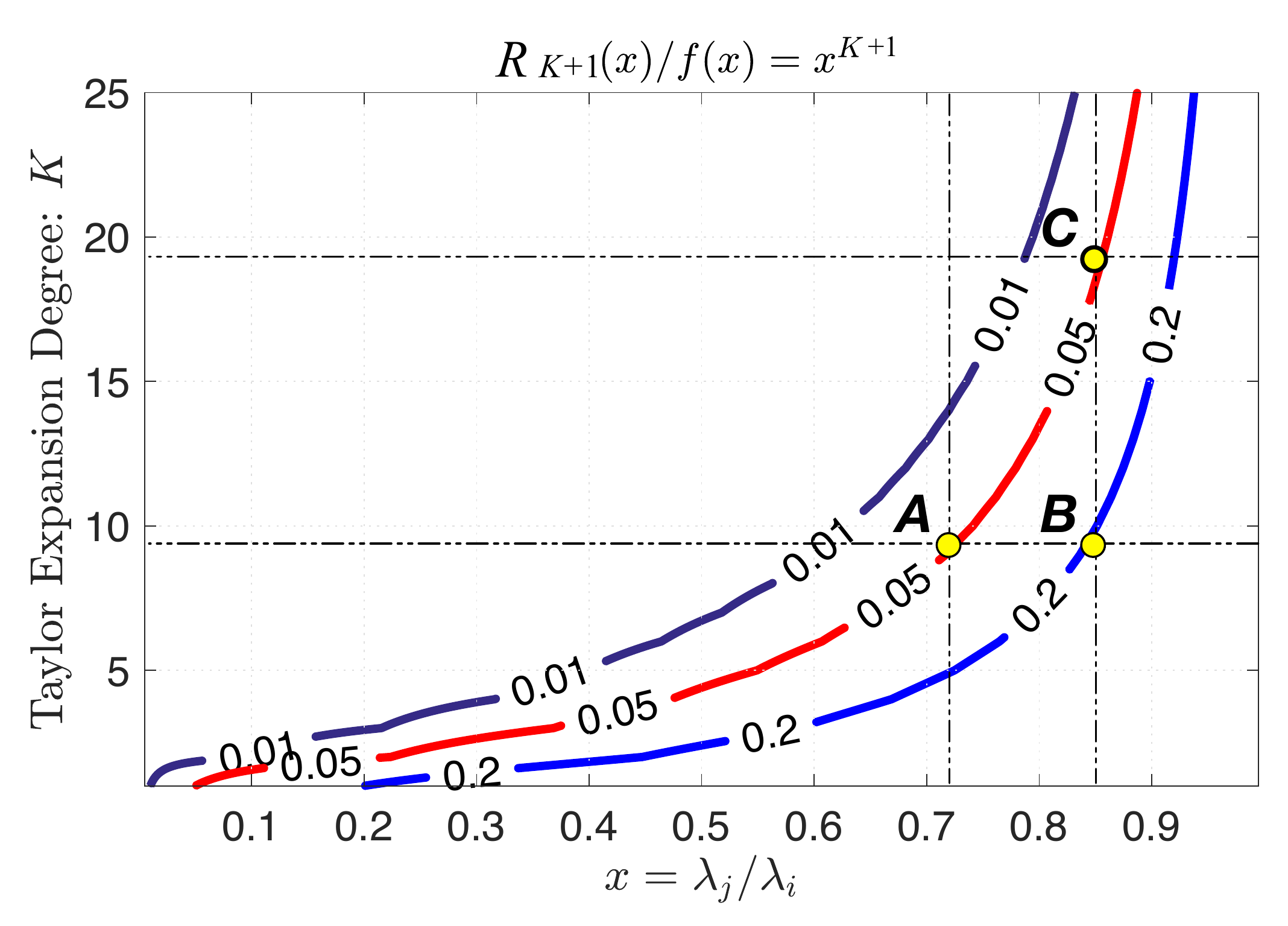} }%
	\caption{{\bf Influence of $K$.} {\bf  (a)} $R_{K+1}(x){=}(\nicefrac{\lambda_k}{\lambda_1})^{K+1}$ as a function of the eigenvalue ratio $\nicefrac{\lambda_k}{\lambda_1}$ and of the Taylor expansion degree $K$. {\bf (b)} Contour plot of the surface shown in (a).}
	\label{fig: ratio}
\end{figure*}

A popular alternative to solving the gradient explosion problem is to clip the gradients. Here, we show that, unlike our Taylor-based approach that yields bounded direction errors of the columns in matrix $\widetilde{\mK}$ which is used to compute the gradients, clipping does not offer any guarantees on this. 

To demonstrate that gradient clipping may yield larger direction errors than Taylor expansion based method, we provide in  Fig.~\ref{fig: clip}  an example that illustrates the direction deviation of the last column of matrix $\widetilde{\mK}$ in dimension 3. Truncating the large value (\ie, $\mK_{23}$) to $\widehat{\mK}_{23}$ by gradient clipping, makes the gradient lean towards the horizontal axis, changing the original direction $[\mK_{13}, \mK_{23}]$  ({\it black arrow}) to $[\mK_{13}, \widehat{\mK}_{23}]$ ({\it blue arrow}).
If, instead, both the small and large values are modified, as with Taylor expansion, the approximate gradient direction remains closer to the original one, as indicated by the {\it red arrow} $[\widehat{\mK}_{13}, \widehat{\mK}_{23}]$.

In this specific example, the problem could of course be circumvented by using a larger threshold which may better preserve descent direction. However, in practice, one will always encounter the cases in which the truncated value will lead to descent direction errors as large as $45^{\circ}$ while our Taylor expansion method will limit the direction error to a much smaller value under the same setup. To check the details, please refer to Section~\ref{sec: grad-clip} in the appendix.

One could argue in favor of scaling the gradient by its own norm, which would keep the descent direction unchanged. However, this is not applicable to the case where two eigenvalues are equals. When $\lambda_i{=}\lambda_j$, $\mK_{i,j} {=} \frac{1}{\lambda_i{-}\lambda_j} \rightarrow \infty$. The norm of the gradient of $K$ then also becomes $\infty$, and the value of $\mK_{i,j}$ after scaling is $\mK_{i,j}/\vert\mK\vert{=}\infty/\infty$, or NaN in our python implementation, which causes training failure.

\begin{figure*}[t!]
	\centering
	\includegraphics[width=0.95\linewidth]{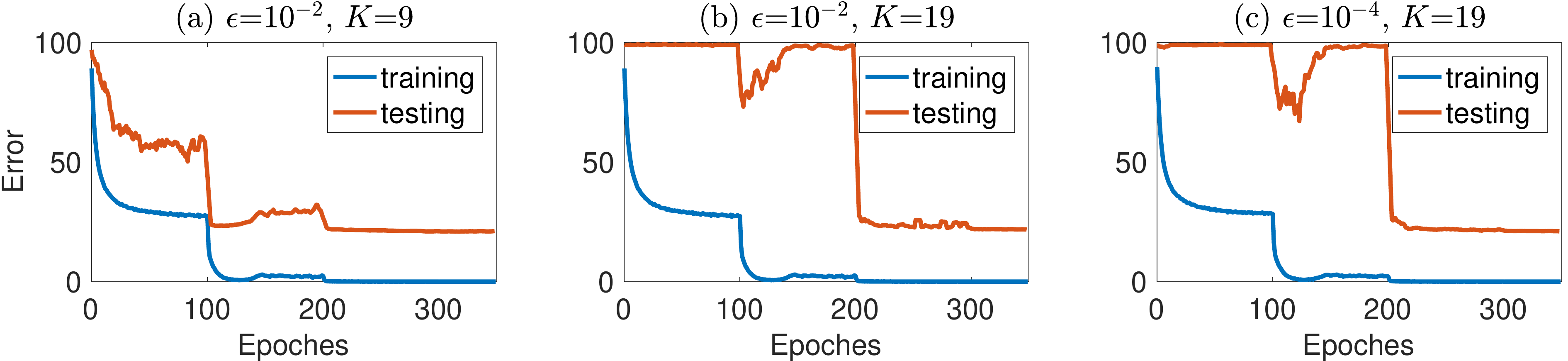}
	\caption{Convergence curves of ResNet18 on CIFAR100 with different hyperparameter values. The matrix dimension $d=64$.}
	\label{fig: hyperparameter}
\end{figure*}

\section{Influence of the Hyperparameters}
\label{sec: hyper}

Our method requires setting two hyperparameters: the Taylor expension degree $K$, and the value $\epsilon$ added to the diagonal of the symmetric matrix. In this section, we study the influence of these hyperparameters on our gradient approximation.

\begin{table*}[t!]
	\caption{ResNet18 \& ResNet50 share the same structure, but the blocks and the FC layer are different. Compared with the original network~\cite{He16a}, we have changed the strides and have replaced the Batch Normalization layer after the first convolutional layer by a {\bf Decorrelated Batch Normalization} layer, which applies ZCA whitening to the feature map.}
	\vspace{-2mm}
	\label{tab: resnet}
	\centering
	\setlength\tabcolsep{6pt}
	\begin{tabular}{ccccccccc}
		\toprule
		Layers & Conv1                  & Norm Layer & Block1   & Block2   & Block3   & Block4   & Average Pooling & FC  \\ 
		\midrule
		Stride  & stride=1,kernel=3$\times$3 & {\bf Decorrelated BN (ZCA-Whitening)} & stride=1 & stride=2 & stride=2 & stride=2 & 4$\times$4       & - \\
		\bottomrule
	\end{tabular}
	\vspace{-2mm}
\end{table*}

\subsection{Hyperparameter $K$}

As discussed in Section~\ref{sec: svd-cov}, when $K$ is small, the remainder of our gradient approximation will be large as more gradient energy will be trimmed off. However, this will also lead to a low upper bound, as can be verified from Eq.~\ref{eq: svd-taylor-form-bound}, and make the gradient less prone to exploding.  
In Figure~\ref{fig: ratio}, we visualize the ratio between the remainder $R_{K+1}(x)$ and the true value $f(x) = \widetilde{\mK}_{i,j}$, as a function of $x = \lambda_j/\lambda_i\in[0,1]$ and of $K$. Note that, as can be verified from Eqs.~\ref{eq: expansion-svd-sym} and~\ref{eq: reminder-ed-talor}, with this notation, this ratio is equal to $x^{K+1}$. In essence, this ratio can be thought of as the proportion of the gradient energy that our approximation discards, and thus, the lower the better.

As can be seen in Fig.~\ref{fig: ratio} (a), the ratio is close to 0 for a large portion of the surface. Let us now look at the cases where it isn't. To this end, we consider the red and blue curves, corresponding to a ratio value of 5\% and 20\%, respectively. Furthermore, let us focus on the points A, B, and C, on these two curves. The contour plot of the two curves are shown in Figure~\ref{fig: ratio}~(b).

For $K{=}9$ a larger value $x{=}\lambda_{j}/\lambda_{i}$ leads to point B, which discards more energy than A. This behavior is beneficial, as it decreases the risk of gradient explosion that is brought about by an increasing $x$ value.

For a large value $x{=}\lambda_{i}/\lambda_{j}{=}0.85$, using a large $K{=}19$ would lead to discarding only 5\% of the energy (point C), but may be less robust than using $K{=}9$, corresponding to B. Therefore, B is a better option and a small $K$ is preferred as it stabilizes the computations, as shown in Figure~\ref{fig: hyperparameter}.

Too small a value $K$, however, will remove too much of the energy, and, as shown by our experiments in Section \ref{subsec: hyper-zca}, we have found $K{=}9$ to be a good tradeoff between stability and accuracy.

\subsection{Hyperparameter $\epsilon$}
The most important function of $\epsilon$ is to avoid having eigenvalues that equal to 0, thus preventing the gradient from exploding, as shown by the upper bound in Eq.~\ref{eq: svd-taylor-form-bound}. Therefore, one might think of favoring large values $\epsilon$, particularly considering that, as shown in Section~\ref{sec: svd-cov}, $\epsilon$ has no influence on the eigenvectors. Nevertheless, too large an $\epsilon$ will make the symmetric matrix deviate far from the original one, thus essentially altering the nature of the problem. In practice,  we typically set $\epsilon \in \{0.01, 0.001, 0.0001\}$.

\begin{table}[!t]
	\caption{ResNet18 on CIFAR100 with matrix dimension $d{=}64$. Hyperparameter values $(\epsilon,K){=}(10^{-2},9)$ yield more stable results and better accuracy.}
	\vspace{-2mm}
	\label{tab: hyper-parameter}
	\setlength{\tabcolsep}{6pt}
	\centering
	\begin{tabular}{@{}llll@{}}
		\toprule
		& Figure.\ref{fig: hyperparameter}(c) & Figure.\ref{fig: hyperparameter}(b) & Figure.\ref{fig: hyperparameter}(a)     \\ 
		\midrule
		($\epsilon,K$)  & ($10^{-4}, 19$) & ($10^{-2}, 19$) & ($10^{-2}, 9$)     \\ 
		Success Rate  & 50.0\%      & 87.5\%      & {\bf 100\%} \\ 
		Min Error  & 21.04      & 21.12      & {\bf 20.91}         \\ 
		Mean Error & 21.31+0.33 & 21.45+0.26 & {\bf 21.14+0.20}    \\ 
		\bottomrule
	\end{tabular}
\end{table}

\section{Experiments}

To demonstrate the stability, scalability, and applicability of our approach, we test it on three different tasks, feature decorrelation~\cite{Huang19}, second order pooling~\cite{Li17a}, and style transfer~\cite{Cho19}. The first two focus on image classification while the third targets a very different application and all three involve performing an SVD on covariance matrices. 

More specifically, we first use feature decorrelation to compare the stability, speed and performance of our new approach against that of our previous one~\cite{Wang19b} and of heuristic gradient clipping method.  To test scalability, we use the second-order pooling task in which the covariance matrix size is $d{=}256$, that is, much larger than the matrix size $d{=}64$ in the feature decorrelation task. Finally, the style transfer task shows that our approach can be applied to other contexts apart from image classification. 

\subsection{Feature Decorrelation}

As a first set of experiments, we use our approach to perform the same task as in~\cite{Wang19b}, that is, image classification with ZCA whitening to generate decorrelated features. ZCA whitening relies on computing a $d\times d$ covariance matrix, with $d\in \{4,8,16,32,64\}$, as in~\cite{Wang19b}. ZCA whitening has been shown to improve classification performance over batch normalization~\cite{Lei18}. We will demonstrate that we can deliver a further boost by making it possible to handle larger covariance matrices within the network. To this end, we compare our approach (SVD-Taylor), which relies on SVD in the forward pass and on the Taylor expansion for backpropagation, with our previous work~\cite{Wang19b} (SVD-PI), which uses PI for backpropagation. We also include the standard baselines that use either SVD or PI in both the forward and backward pass, denoted as SVD and PI, respectively. 

\subsubsection*{Network structure}

In all the experiments of this section, we use either Resnet18 or Resnet50~\cite{He16a} as our backbone. As shown in Table~\ref{tab: resnet}, we retain their original architectures but introduce an additional layer between the first convolutional layer and the first pooling layer. The new layer computes the covariance matrix of the feature vectors, performs SVD, and uses the eigenvalues and eigenvectors as described in Alg.~\ref{alg: svd-taylor}.
To accommodate this additional processing, we change the stride $s$ and kernel sizes in the subsequent blocks as shown in Table~\ref{tab: resnet}.

\subsubsection*{Training Protocol}

We evaluate our method on the CIFAR10, CIFAR100, and Tiny ImageNet datasets~\cite{Krizhevsky09,TinyImageNet}. On CIFAR10 and CIFAR100, we use 350 stochastic gradient epochs to train the network with a learning rate of 0.1, a momentum=$0.9$, a weight decay=$5{\times}10^{-4}$, and a batch size=$128$. We decay the initial learning rate by $\gamma{=}0.1$ every 100 epochs. On Tiny ImageNet, following common practice~\cite{Li18i}, we train for 90 epochs, decaying the initial learning rate by $\gamma{=}0.1$ every 30 epochs, and use the same values as in our other experiments for the remaining hyperparameters.

In the remainder of this section, we first study the influence of our hyperparameters in this context to find their appropriate values, and, unless otherwise specified, use these values by default when backprogating the gradients in the following experiments. We then compare our results to the baselines discussed above. 

\begin{figure*}[t!]
	\centering
	\subfloat[$d{=}4$]{\includegraphics[width=0.333\linewidth]{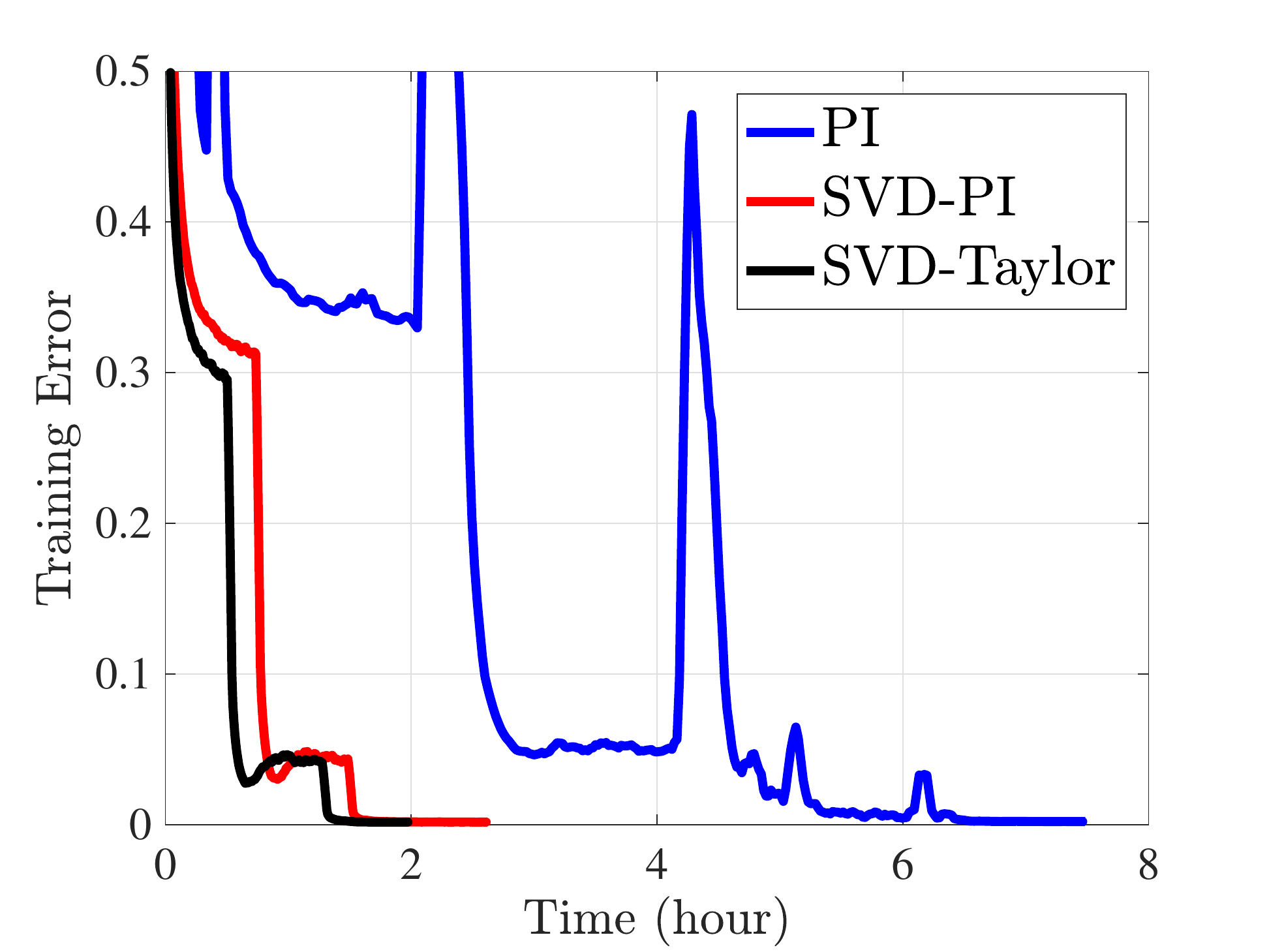} }%
	\subfloat[$d{=}8$]{\includegraphics[width=0.333\linewidth]{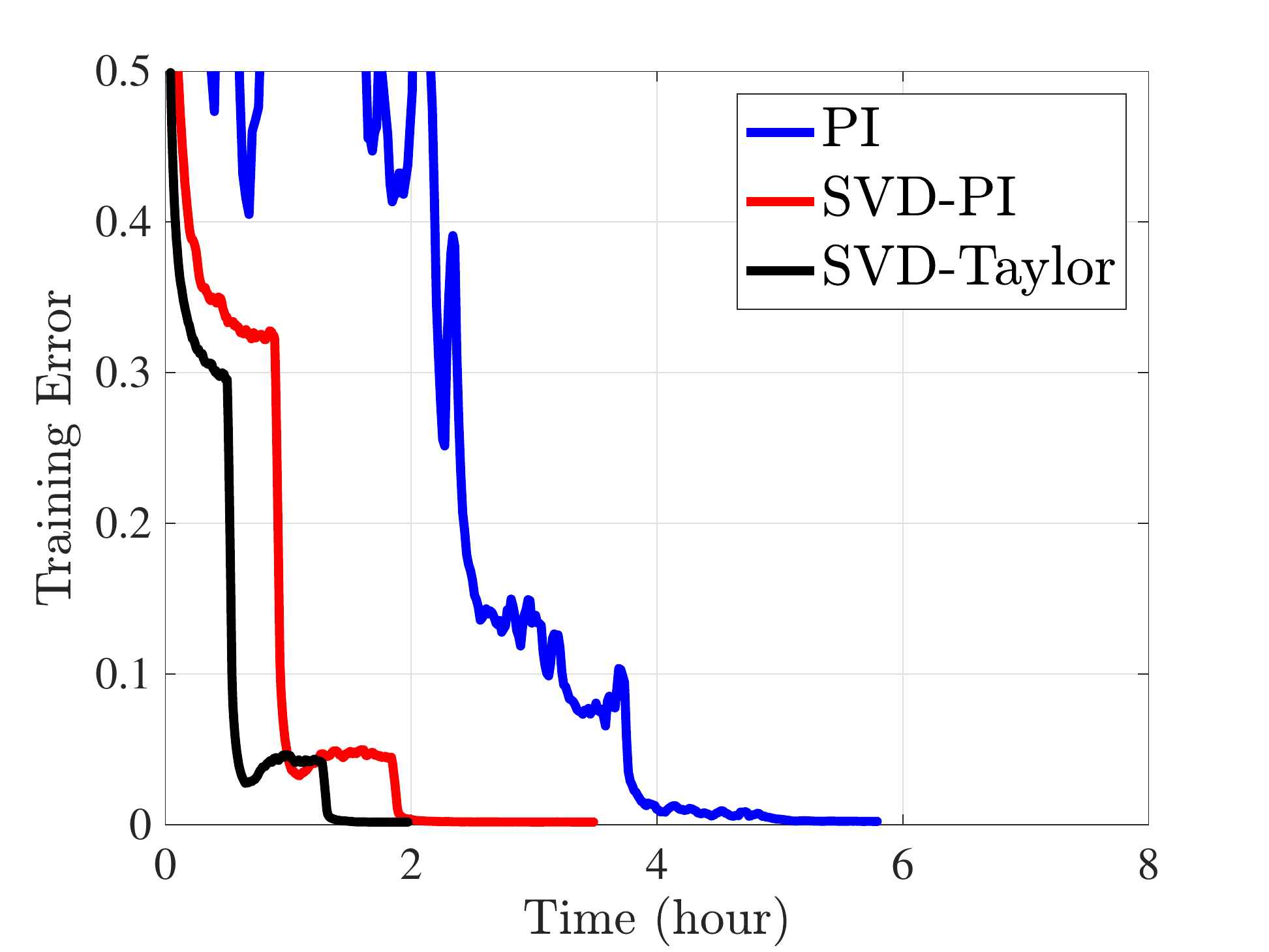} }%
	\subfloat[$d{=}16$]{\includegraphics[width=0.333\linewidth]{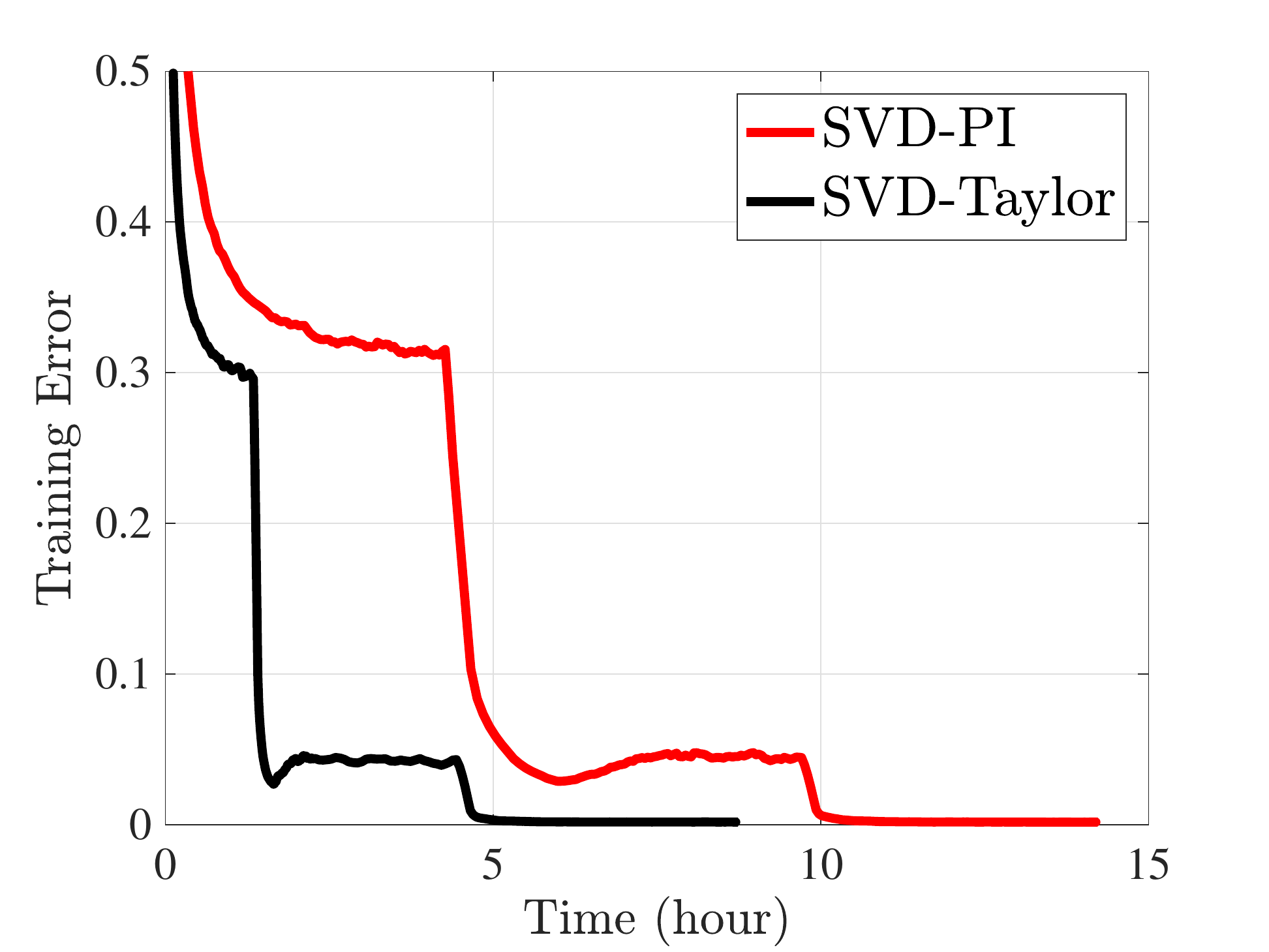} }%
	\caption{Convergence curves for different matrix dimensions $d$ of our Taylor-based method (denoted as SVD-Taylor), the PI-based one of~\cite{Wang19b} (denoted as SVD-PI), and the standard power iteration method~\cite{Mang17}. Both SVD-Taylor and SVD-PI use standard SVD in the forward pass.
		Since our approach yields better gradients, it converges faster at the beginning. Furthermore, although all methods converge to the same training error, our approach yields better testing accuracy. All methods are trained with 350 epochs. The learning rate is decreased by 0.1 every 100 epochs. The x axis represents the time consumed and y axis denotes the training error.}
	\label{fig: convergence-speed}
\end{figure*}

\subsubsection{Hyperparameters}
\label{subsec: hyper-zca}

To illustrate the discussion of Section~\ref{sec: hyper}, we evaluate our approach with different hyperparameter settings on CIFAR100~\cite{Krizhevsky09} with a ResNet18 network. To this end, in addition to the classification error, we report the success rate of our method, corresponding to the percentage of the time it converges without undergoing gradient explosion. 
We run each setting 8 times to compute this success rate.
This metric is directly related to the hyperparameters because of their effect on the gradient upper bound.

As shown in Table~\ref{tab: hyper-parameter}, for $K{=}19$, increasing $\epsilon$ from $10^{-4}$ to $10^{-2}$ yields a success rate increase from 50\% to 87.5\%. After decreasing $K$ from 19 to 9, we do not observe any failure cases, and the classification accuracy is the best among the 3 settings studied here, because of the resulting healthier gradients. 

In Figure~\ref{fig: hyperparameter}, we plot the training and testing error curves for the same 3 hyperparameter settings. These curves further illustrate the instabilities that can occur during training with too large values of $K$, such as $K{=}19$. In particular, while the training error decreases reasonably smoothly, the testing one is much less stable and inconsistent with the training error.
This is due to the fact that, with a large matrix dimension $d{=}64$, many eigenvalues go to zero, or rather $\epsilon$. 
Specifically, in this experiment, we observed on average 20 eigenvalues equal to $\epsilon$ during the first training epoch. 
Nevertheless, with $K{=}9$, training proceeds more smoothly, and the test error better follows the training one. In the following experiments, we therefore use $K{=}9$ and $\epsilon {=} 0.01$.

\begin{table*}[t!]
	\caption{Errors and success rates of ResNet18 with standard SVD, Power Iteration (PI), SVD-PI \& our SVD-Taylor on CIFAR10 with the image size of $32{\times}32$. $d$ is the matrix size of each feature group we process individually.}
	\label{tab:cifar10}
	\centering
	\setlength\tabcolsep{12pt}
	\begin{tabular}{lllllll}
		\hline\noalign{\smallskip}
		Methods & Error & $d=4$       & $d=8$       & $d=16$      & $d=32$      & $d=64$      \\ \midrule
		\multirow{3}{*}{SVD}    
		& Success Rate           & 46.7\%    & 0\%       & 0\%       & 0\%       & 0\%       \\ 
		& Min                    & 4.59      & -         & -         & -         & -         \\ 
		& Mean                   & $4.54\pm0.08$ & -         & -         & -         & -         \\ 
		\midrule
		\multirow{3}{*}{PI}      
		& Success Rate           & \textbf{100\%}     & 6.7\%     & 0\%       & 0\%       & 0\%       \\ 
		& Min                    & 4.44      & 6.28      & -         & -         & -         \\ 
		& Mean                   & 4.99$\pm$0.51 & -         & -         & -         & -         \\ 
		\midrule
		\multirow{4}{*}{SVD-Clip}
		& Success Rate           & \textbf{100\%}     & \textbf{100\% }    & \textbf{100\%}     & \textbf{100\%}     & \textbf{100\%}     \\ 
		& Min                    & 4.57      & 4.66      & 4.70      & 4.76      & 4.62      \\ 
		& Mean                   & 4.84$\pm$0.23 & 4.77$\pm$0.66 & 4.89$\pm$0.16 & 4.87$\pm$0.13 & 4.83$\pm$0.15 \\ 
		\midrule
		\multirow{4}{*}{SVD-PI~\cite{Wang19b}  }
		& Success Rate           & \textbf{100\%}     & \textbf{100\% }    & \textbf{100\%}     & \textbf{100\%}     & \textbf{100\%}     \\ 
		& Min                    & 4.59      & 4.43      & 4.40      & 4.46      & 4.44      \\ 
		& Mean                   & 4.71$\pm$0.11 & 4.62$\pm$0.18 & 4.63$\pm$0.14 & 4.64$\pm$0.15 & 4.59$\pm$0.09 \\ 
		\midrule
		\multirow{4}{*}{SVD-Taylor}
		& Success Rate           & \textbf{100\% }    & \textbf{100\%}     & \textbf{100\%}     & \textbf{100\%}     & \textbf{100\% }    \\ 
		& Min           & \textbf{4.33}     & 4.34      &\textbf{4.33}      & 4.40      & 4.40     \\
		& Mean                   & 4.52$\pm$0.09 & 4.55$\pm$0.12 & 4.52$\pm$0.14& 4.57$\pm$0.16 & {\textbf{4.50$\pm$0.08}} \\  
		\bottomrule
	\end{tabular}
\end{table*}

\begin{table*}[t!]
	\caption{Error rates of ResNet18/50 with our ZCA layer on CIFAR100 dataset with the image size is $32{\times}32$. We compare our previous work SVD-PI~\cite{Wang19b} with our current one SVD-Taylor. $d$ is the size of the feature groups we process individually.}
	\label{tab:cifar100}
	\centering
	\setlength\tabcolsep{8pt}
	\begin{tabular}{@{}llllllll@{}}
		\toprule
		Methods                               & Error & d=4        & d=8        & d=16       & d=32       & d=64    & {\bf Batch Norm}   \\ \midrule
		\multirow{2}{*}{ResNet18~SVD-PI~\cite{Wang19b}}      & Min   & 21.03      & 21.15      & 21.14      & 21.36      & 21.04  & 21.68    \\ 
		& Mean  & 21.51$\pm$0.28 & 21.56$\pm$0.35 & 21.45$\pm$0.25 & 21.58$\pm$0.27 & 21.39$\pm$0.23 & 21.85$\pm$0.14 \\
		\multirow{2}{*}{RestNet18~SVD-Taylor} & Min   & 20.99  &    20.88 &   20.86   &    20.71   &  20.91    \\
		& Mean  &   {\bf 21.24$\pm$0.17}  & {\bf 21.32$\pm$0.31}  &   {\bf 21.30$\pm$0.33}  &    {\bf 20.99$\pm$0.27}  &    {\bf 21.14$\pm$0.20}   \\
		\midrule
		\multirow{2}{*}{ResNet50~SVD-PI~\cite{Wang19b}}      & Min   & 20.66      & 20.15      & 19.78      & 19.24      & 19.28   &20.79   \\ 
		& Mean  & 20.98$\pm$0.31 & 20.59$\pm$0.58 & 19.92$\pm$0.12 & 19.54$\pm$0.23 & 19.94$\pm$0.44 & 21.62$\pm$0.65\\ 
		\multirow{2}{*}{ResNet50~SVD-Taylor}  & Min   &  19.55  &   19.21   & 19.36  &    19.15 &  19.26   \\
		& Mean  &   {\bf 20.34$\pm$0.53}  &  {\bf 19.57$\pm$0.24}  &   {\bf 19.60$\pm$0.19}   &  {\bf 19.47$\pm$0.24 } &  {\bf 19.81$\pm$0.24}  \\ \bottomrule
	\end{tabular}
\end{table*}

\begin{table*}[t!]
	\caption{Error rates of ResNet18/50 with our ZCA layer on Tiny ImageNet \textit{val.} set with the image size of $64{\times}64$. $d$ is the size of the feature groups we process individually.}
	\label{tab:tinyimgnet}
	\centering
	\setlength\tabcolsep{8pt}
	\begin{tabular}{@{}llllllll@{}}
		\toprule
		Methods                               & Error & d=4        & d=8        & d=16       & d=32       & d=64    & {\bf Batch Norm}   \\ \midrule
		\multirow{2}{*}{ResNet18~SVD-PI~\cite{Wang19b}}      & Min   &   36.24    & 36.78      & 36.88      & 36.78      & 36.48  & 36.92    \\ 
		& Mean  & 36.62$\pm$0.46 & 38.43$\pm$1.66 & 37.27$\pm$0.36 & 38.43$\pm$1.65 & 36.83$\pm$0.22 & 37.15$\pm$0.21\\
		\multirow{2}{*}{RestNet18~SVD-Taylor} & Min   & 36.30  &  36.42 &   35.68   &    36.96   &  36.16    & \\
		& Mean  &  {\bf 36.58$\pm$0.30}  & {\bf 36.70$\pm$0.29}  &   {\bf 36.64$\pm$0.66}  &    {\bf 37.32-$\pm$0.28}  &    {\bf 36.72$\pm$0.41}   &   \\
		\midrule
		ResNet50~SVD-PI~\cite{Wang19b} & -
		& 35.74 & 35.60 & 35.36 & 35.34 & 35.10 & 35.98\\ 
		ResNet50~SVD-Taylor  & -
		& {\bf 35.18} & {\bf 35.40} & {\bf 35.16} & {\bf 35.02} & {\bf 34.74} & \\ \bottomrule
	\end{tabular}
\end{table*}

\subsubsection*{Why do $K$, $\epsilon$ only affect the test error but not the training one?}

In the training phase, before we compute the Taylor expansion, the eigenvalues $\lambda_i$ of matrix $\mM$ that are smaller than $\epsilon$ are all clamped to $\epsilon$, so as to avoid the explosion of $\nicefrac{1}{\lambda_{i}}$ in the gradient. While this results in inaccuracies in the whitening matrix $\mS$, it has no influence on the training batch. For instance, when $\lambda_i{=}0$ is clamped to $\epsilon$,
the whitening matrix contains terms of the form
\begin{equation}
	\vv_i \lambda_i \vv^T_i \rightarrow \vv_i \epsilon \vv^T_i \neq 0, \quad \text{with} \quad \lambda_i \rightarrow \epsilon.
\end{equation}
However, when applied to the covariance matrix $\mM$, since the $\vv_i$s are orthogonal to each other (\ie $\vv^T_i \vv_j{=}0, i{\neq} j$), and $\mM=\sum_{j=1}^n\vv^T_j \lambda_j \vv_j$, we have
\begin{equation}
	\begin{aligned}
		&\vv_i \epsilon \vv^T_i \mM {=} \vv_i \epsilon \vv^T_i \sum_{j=1}^n\vv_j \lambda_j \vv^T_j {=}\sum_{j=1}^n \vv_i \epsilon \vv^T_i \vv_j \lambda_j \vv^T_j \\
		= & \vv_i \epsilon \vv^T_i \vv_i \lambda_i \vv^T_i = \vv_i \epsilon \lambda_i \vv^T_i = 0, \quad \text{as } \lambda_i {=}0.
	\end{aligned}
\end{equation}
which has 0 gradient. Therefore, $\vv_i \epsilon \vv^T_i$ has no influence on the training batch, but this inaccuracy will be kept in the running subspace $E_S$ used in Alg.~\ref{alg: svd-taylor}. In the testing phase, before training converges, the matrix $\widehat{\mM}$ obtained for an arbitrary testing batch is quite different from $\mM$ and will yield $\vv_i \lambda_i \vv^T_i \widehat{\mM} {\neq} 0$, which will create inconsistencies between training and testing.

\begin{figure}[!htb]
	\centering
	\includegraphics[width=0.75\linewidth]{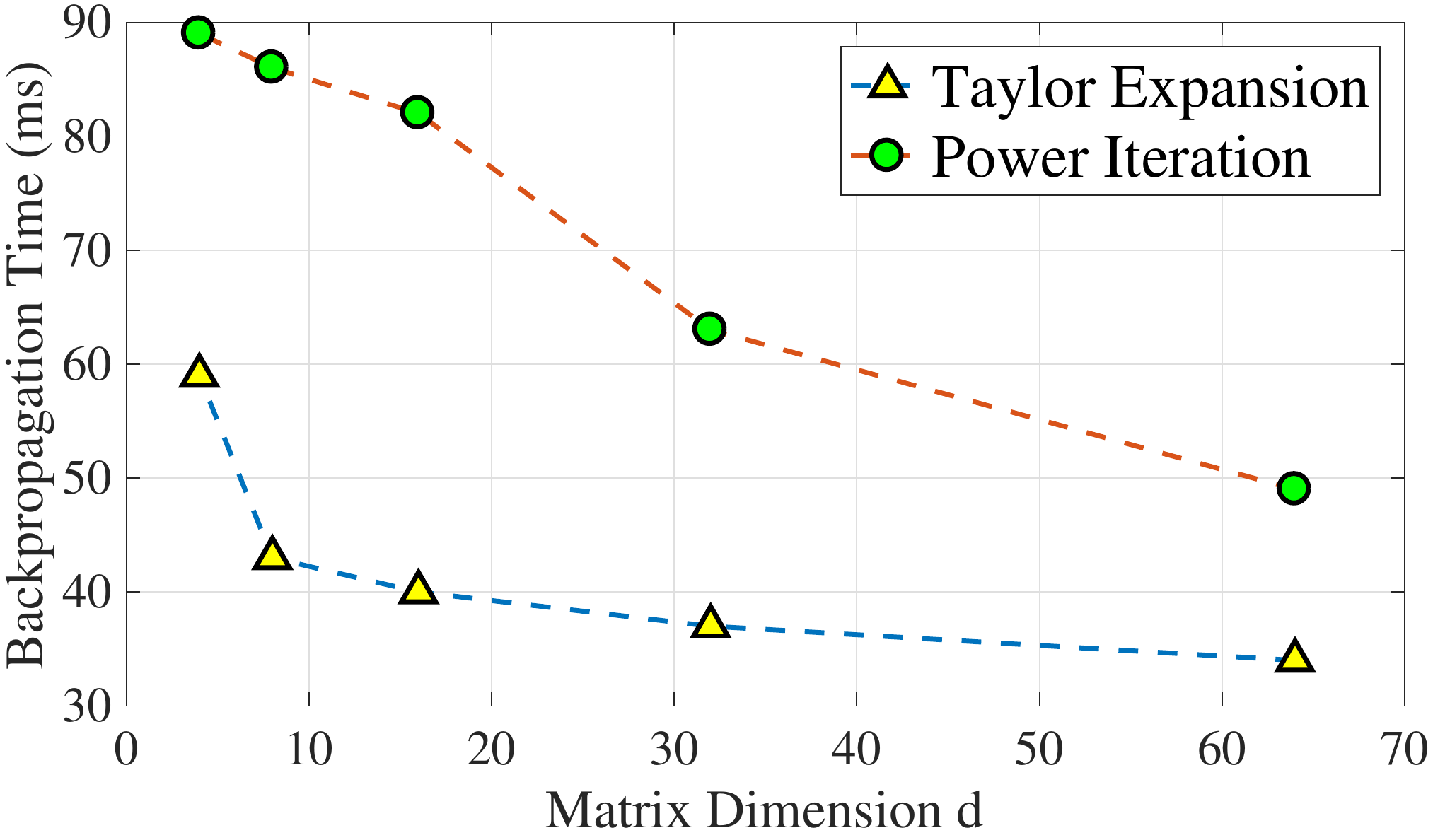}
	\caption{Computation speed: Backpropogation time for ResNet18 on CIFAR10 with different ZCA layers and different matrix dimensions. Taylor expansion is always faster than PI. For fair comparison, the Power Iteration number and the Taylor expansion degree are set to the same value 9. The computation time is averaged over the first 10 epochs. The batch size is 128. This experiment was performed on an Nvidia V100 GPU.}
	\label{fig: speed}
\end{figure}

\subsubsection{Speed and Performance}
\label{sec:speed}

We now turn to the comparison of our approach (SVD-Taylor) against the three baselines introduced above SVD, PI, and SVD-PI~\cite{Wang19b}. The pseudo-code for ZCA whitening is given in Alg.~\ref{alg: svd-taylor}. The algorithm takes as input a ${d \times n}$ matrix $\mX$, where $d$ represents the feature dimensionality and $n$ the number of samples. It relies on eigenvalues and eigenvectors to compute a ${d \times d}$ matrix $\mS$ such that the covariance matrix of $\mS \mX$ is the ${d \times d}$ identity matrix, meaning that the transformed features are decorrelated. ZCA has shown great potential to boost the classification accuracy of deep networks~\cite{Lei18}, but only when $d$ can be kept small enough to prevent the training procedure from diverging. To this end, the $c$ output channels of a convolutional layer are partitioned into $G$ groups so that each one contains only $d=c/G$ features. ZCA whitening is then performed within each group independently. This can be understood as a block-diagonal approximation of the complete ZCA whitening. In~\cite{Lei18}, $d$ is taken to be 3, and the resulting $3 {\times} 3$ covariance matrices are then less likely to have similar or zero eigenvalues. Here, thanks to our more robust gradient computation, we can consider larger values of $d$.
We summarize our empirical findings below.

\subsubsection*{Improved Speed}

In Figure~\ref{fig: speed}, we plot the backpropagation speed when computing the gradient using PI as in~\cite{Wang19b}, or using our Taylor expansion strategy. Note that our approach is consistently faster than the PI-based one, because it does not involve an iterative procedure. Note that speed increases as the matrix dimension increases, which may seem counter-intuitive. This, however, is due to the fact that for $d<64$, we grouped the feature dimensions into several matrices, \eg for $d=4$, we have 16 covariance matrices, whereas for $d=64$, we have a single one. The fact that we process each matrix independently then explains the trend of the curves.

Since our SVD-Taylor method yields better gradients than SVD-PI, it also helps the network to converge faster during training. 
We plot the training losses when using standard PI,  SVD-PI and our SVD-Taylor method as a function of the number of epochs on the CIFAR-10 dataset. 

As shown in Figure~\ref{fig: convergence-speed} (a)(b)(c), the loss curves of our Taylor-based method always decrease faster than that of SVD-PI.
Note also that our SVD-Taylor and SVD-PI are much smoother than the standard PI.
As the standard PI always fails when the dimension is 16, we could not plot its training curve in (c).

\subsubsection*{Improved Performance}

In Table~\ref{tab:cifar10}, we report the results of all the methods, including gradient clipping (SVD-Clip), on CIFAR-10 for different number of groups $G$, corresponding to different matrix dimensions $d$. Specifically, for each method, we report the mean classification error rate over the trials in which it succeeded, along with the corresponding standard deviation and its success rate.
Because of numerical instability, the training process of these methods often crashes. When it comes to success rate, we observed that
\begin{enumerate}[leftmargin=*]
	\item For SVD, when the matrix dimension $d=4$, 8 out of 15 trials failed; when $d{\geq}8$, the algorithm failed every time.
	\item For PI, when the matrix dimension $d=4$, all the trials succeeded;  when $d=8$, only 1 out of 15 trials succeeded; when $d{\geq}16$, the algorithm failed every time.
	\item For both our SVD-Taylor and SVD-PI~\cite{Wang19b}, we never saw a training failure case, independently of the matrix dimension ranging from $d=4$ to $d=64$.
\end{enumerate}

In Table~\ref{tab:cifar10}, our approach consistently outperforms all the baselines. This is not only true for SVD and PI, which have very low success rates as soon as $d>4$, but also for the SVD-Clip and the state-of-the-art SVD-PI~\cite{Wang19b} that was explicitly proposed as a solution to instabilities. This evidences the benefits of using the Taylor expansion to avoid the approximation errors due to the deflation process of PI.
Our approach delivers the smallest error rate for $d=16$, and the smallest mean error for $d=64$, which confirms that increasing the size of the groups can boost performance.
Furthermore, gradient clipping (with a clipping threshold of 100) always yields inferior performance than both SVD-PI~\cite{Wang19b} and our approach, regardless of the matrix dimension. This is because it does not preserve the gradient direction as well as our approach.

We report similar results  on CIFAR-100 in Table~\ref{tab:cifar100}, using either ResNet18 or ResNet50 as the backbone.  Our approach again systematically delivers better performance. Specifically, being able to increase $d$ to 32  for ResNet18 and ResNet50 allows us to outperform batch normalization in terms of min and mean error.

We report our Tiny ImageNet results in Table~\ref{tab:tinyimgnet}. The image size is $64{\times}64$. For ResNet18, we ran each setup 5 times and computed the average performance. Our approach systematically outperforms PI. Because it is very time consuming, we only ran the experiment for ResNet50 once for each setup; the conclusions are the same as for ResNet18.

\subsubsection*{Limitations of Taylor Expansion} 
Our Taylor expansion approach is faster than PI during backpropagation. However, during the forward pass, it consumes the same amount of time, which may be 2 times (with matrix dimension $d=4, 8$) to 10 times (with matrix dimension $d=16, 32$) more than in the backward pass. Therefore, the bottleneck of the SVD layer is the computation speed in the forward pass, and thus designing more efficient ways to compute the eigenvectors constitutes an interesting direction for future research.

\subsection{Second-order Pooling}

In our second set of experiments, we used the second-order pooling task~\cite{Ionescu15} on the large scale ImageNet dataset~\cite{Russakovsky15} to compare the stability of SVD-PI, our earlier approach~\cite{Wang19b}, against that of SVD-Taylor, the Taylor-based one we advocate here. This task involves covariance pooling of the deep features rather than traditional first-order pooling, such as max or average pooling. Different from the previous feature decorrelation task in which the maximum size of the covariance matrix is set to $d{=}64$, the covariance matrix size in this new task is set to $d{=}256$, which is more challenging and allows us to check whether either method scales up to large matrices.

The covariance matrices are computed as the global image representations and used to boost the performance of large-scale classification. This requires an SVD of the covariance matrix of the feature maps. As for the decorrelated batch normalization task~\cite{Huang19}, the covariance matrix characterizes the correlations of feature channels.

\subsubsection{Method}

\setlength{\textfloatsep}{0pt}
\begin{algorithm}[t!]
	\caption{Second-order Pooling with SVD-Taylor}
	\label{alg: svd-second-order}
	{Centralize $\mX$: $\mu{\leftarrow}\bar{\mX}$, $\widetilde{\mX}{\leftarrow}\mX{-}\mu\bf{1}^{\top}$}\;
	{Compute Covariance Matrix: $\mM{\leftarrow}\widetilde{\mX}\widetilde{\mX}^{\top}{+}\epsilon I$}\;
	{{\bf Forward pass}: }
	{Standard SVD: $\mV{\Lambda}\mV^{\top}{\leftarrow}\textrm{SVD}(\mM)$; $\lambda_i{\leftarrow}\max(\lambda_i, \epsilon), i{=}1,2,\cdots, n$ $\Lambda_{diag} {\leftarrow} [\lambda_1, \cdots, \lambda_n],\; \mV{\leftarrow}\left[\vv_1, \cdots, \vv_n\right]$}\;
	Compute covariance pooling transformation matrix: $\mS \leftarrow \mV f(\Lambda)\mV^{\top}$\;
	Second-order Pooling:
	$\mX \leftarrow \mS\widetilde{\mX}$\;
	{{\bf Backward pass}: }
	{Compute the Taylor expansion of $\widetilde{\mK}$ according to Eq.~\ref{eq: expansion-svd-sym}}\;
	{Compute the gradient using Eq.~\ref{eq: svd-cov}.}
\end{algorithm}

The second-order pooling algorithm is shown in Alg.~\ref{alg: svd-second-order}.
The main difference between decorrelated batch normalization and second-order pooling lies in how the transformation matrix $\mS$ is defined. In both cases, the transformation can be written as
\begin{equation}
	\mS \leftarrow \mV \; f(\widetilde{\Lambda}) \; \mV^{\top} \; ,
\end{equation}
where $f(\widetilde{\Lambda})$ is a function of the eigenvalues~\cite{Li17a}. For decorrelated batch normalization, it is simply set to $\lambda_{i}^{-\frac{1}{2}}$, whereas, for  second-order pooling, it is taken to be $f(\widetilde{\Lambda}) = \text{diag} (f(\lambda_1), f(\lambda_2), \cdots, f(\lambda_n))$, where $f(\lambda_i)$ is defined as the power of eigenvalues $f(\lambda_i) = \lambda_{i}^{\alpha}$ with $\alpha$ set to $\frac{1}{2}$. For improved performance, one can also normalize these values by taking them to be
\begin{equation}
	f(\lambda_i) = \frac{\lambda_{i}^{\alpha}}{\sqrt{\Sigma_k \lambda_k^{2\alpha}}} \;.
\end{equation}
The second-order pooling layer is usually located at the end of the network to get better features for the classifier. By contrast, the feature decorrelation layer is usually put in the beginning of the network to get decorrelated features to help the following convolutional layers to get better features. Moreover, the second-order pooling layer does not store the statistics of the running mean or covariance matrix as the feature decorrelation layer.

For the second-order pooling task, we used the same training setup as in~\cite{Li17a} with AlexNet~\cite{Krizhevsky12} as our backbone but we replaced the SVD gradients by ours. Because of our limited computational resources, we  ran the experiment only once and report its results in Table~\ref{tab:imgnet}. Note that in~\cite{Li17a}, the gradients are computed as in~\cite{Ionescu15} and the truncation technique of Section~\ref{sec: truncation} is used to prevent gradient explosion.

\subsubsection{Stability and Performance}

\begin{table}[t!]
	\caption{Error rate (\%) of covariance pooling methods with AlexNet on ImageNet \textit{val.} set.}
	\label{tab:imgnet}
	\vspace{-3mm}
	\centering
	\setlength\tabcolsep{8pt}
	\begin{tabular}{@{}lcc@{}}
		\toprule
		Methods                               & Top-1 Error        & Top-5 Error   \\ \midrule
		AlexNet~\cite{Krizhevsky12} & 41.8 &19.2 \\
		\midrule
		Bilinear CNN~\cite{Lin15} & 39.89 & 18.32 \\
		Improved Bilinear CNN~\cite{Lin17a} & 40.75 & 18.91 \\
		DeepO{\tiny 2}P\cite{Ionescu15} & 42.16 & 19.62\\
		G2DeNet~\cite{Wang17b} & 38.71 & 17.66 \\
		\midrule
		MPN-COV~\cite{Li17a} & 38.51 & 17.60 \\
		Fast MPN-COV (\ie iSQRT-COV)~\cite{Li18i} & {\bf 38.45} & 17.52 \\
		SVD-PI~\cite{Wang19b} & - & - \\
		SVD-Taylor(ours) & 38.74 & {\bf 17.12} \\
		\bottomrule
	\end{tabular}
\end{table}

From Table~\ref{tab:imgnet}, we can observe that for the large scale dataset (\ie ImageNet) whose image size ($256 {\times} 256$) is much larger than CIFAR ($32{\times}32$) and Tiny ImageNet ($64{\times}64$), our earlier SVD-PI~\cite{Wang19b} failed to converge.
Actually, for large images, the covariance matrices are also very large. As a consequence, many eigenvectors correspond to very small eigenvalues. However, the SVD-PI~\cite{Wang19b} approach discards these eigenvectors with small eigenvalues even though they contain useful information, as shown from Line 8 to Line 13 in Alg.~\ref{alg: svd-pi}. By contrast our Taylor-SVD preserves all the eigenvectors and thus converges. As shown in Table~\ref{tab:imgnet}, our new approach has comparable performance with the state-of-the-art method in terms of top-1 error and yields the best top-5 error.

\subsection{Style Transfer}
\label{sec: style-transfer}

\begin{figure*}[t!]
	\includegraphics[width=\linewidth]{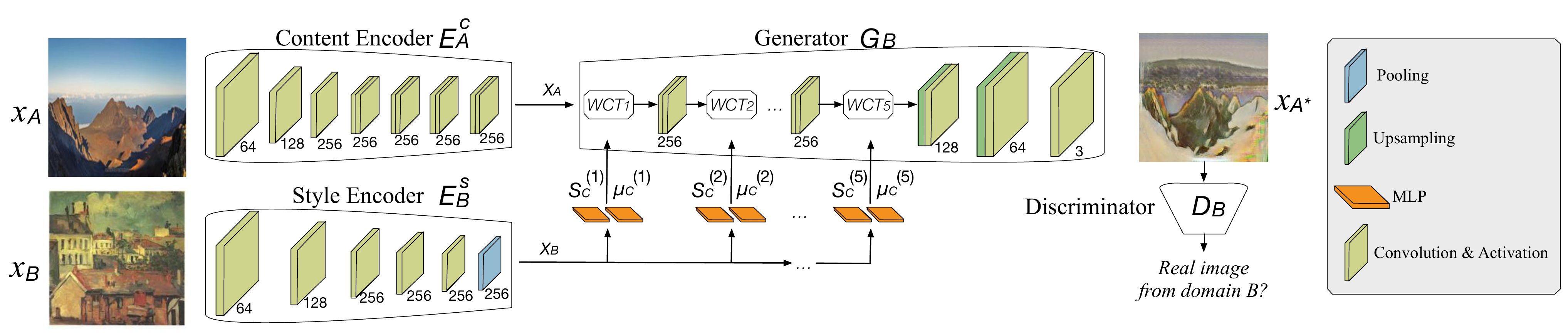}
	\vspace{-4mm}
	\caption{Network architecture to transfer the style from an image $x_B$ to a content image $x_A$. The content encoder $E_A^C$ extracts content features $\mX_\mA$ from $x_A$, while the style encoder $E_B^S$ computes style features $\mX_\mB$ from $x_B$. 
		Following~\cite{Cho19}, the style features are then fed to 5 pairs of multilayer perceptron networks (MLP), resulting in 5 coloring transformation matrices $(\mS_c^{(1)}, \mu_c^{(1)})$, $(\mS_c^{(2)}, \mu_c^{(2)})$, $\cdots$, $(\mS_c^{(5)}, \mu_c^{(5)})$ ranging from high-level to low-level features.
		WCT is then performed to transfer the style from $\mX_\mB$ to the content of $\mX_\mA$ in the generator $G_B$, and finally obtain the stylized image $x_{A^*}$.
		The discriminator $D_B$, used in an adversarial fashion, helps to improve the realism of $x_{A^*}$ by discriminating between real images from domain $B$ and generated images.}
	\label{fig: wct-pip}
	\vspace{-4mm}
\end{figure*}

In our final set of experiments, we use of our approach for style transfer. Existing image style transfer techniques~\cite{Chen17e,Johnson16,Ulyanov16b,Lee18b} commonly exploit higher-order statistics of the learned features, such as their covariance matrix, to represent the style~\cite{Huang18c,Huang17e}. Below, we first discuss the formulation of this approach and the network architecture used in our experiments, and finally present our results.

\subsubsection{Whitening \& Coloring Transformation (WCT)}
Style transfer is typically achieved by applying whitening and coloring transformations~\cite{Cho19,Chiu19} to some feature representation of an image. When WCT is directly applied to the raw RGB pixel values, it only transfer the colors. This is illustrated in Figure~\ref{fig: zca-pca} (d) (e), where the colors of the style images B are transferred to content images A.  Nevertheless, WCT can be used with other representations, such as the features extracted by a backbone network. Let us now briefly review the details of these two transformations.

\subsubsection*{Whitening Transformation}
The whitening transformation closely resembles the ZCA operation described in Alg.~\ref{alg: svd-taylor}.  Specifically, given a feature matrix $\mX_\mA{\in}\R^{C{\times}BHW}$ extracted from a source content image, with $(C,B,H,W)$ denoting the number of channels, batch size, feature map height and width, we first subtract the mean of each channel in $\mX_\mA$. Applying a whitening transformation to such a zero-mean matrix then changes its covariance matrix to the identity matrix, which removes its style. To achieve this, we rely on the transformation matrix
\begin{equation}
	\mS_w {=} \mV_w (\Lambda_w)^{-\frac{1}{2}}\mV_w^{\top}\;,
\end{equation}
where $\mV_w$ and $\Lambda_w$ are the matrices of eigenvectors and eigenvalues of the covariance matrix of $\mX_s$.
The whitened features are then computed as
\begin{equation}
	\mX_w {=} \mS_w (\mX_\mA-\mu_w)\;,
\end{equation}
where $\mu_w$ is a matrix replicating $BHW$ times the $C$-dimensional channel-wise mean vector of $\mX_\mA$.

\subsubsection*{Coloring Transformation}
Once the content features have been whitened, we aim to re-color them with the style of a target style feature map $\mX_\mB$. In essence, this is achieved by the inverse process of the whitening operation. That is, we compute a coloring transformation matrix
\begin{equation}
	\mS_c {=} \mV_c (\Lambda_c)^{\frac{1}{2}}\mV_c^{\top}\;,
\end{equation}
where $\mV_c$ and $\Lambda_c$ are the matrices of eigenvectors and eigenvalues of the covariance matrix of $\mX_{\mB}$.
Note that the power of $\Lambda_c$ in $\mS_c$ is $\nicefrac{1}{2}$, not $-\nicefrac{1}{2}$ as in the whitening transformation matrix $\mS_w$. The colored features with the new style are then obtained as
\begin{equation}
	\mX_{\mA^{*}} {=} \mS_c \mX_w+\mu_c\;,
\end{equation} 
where $\mu_c$ is a similar matrix to $\mu_w$, but containing the channel-wise means of $\mX_\mB$.

\begin{figure*}[t!]
	\centering
	\includegraphics[width=\linewidth]{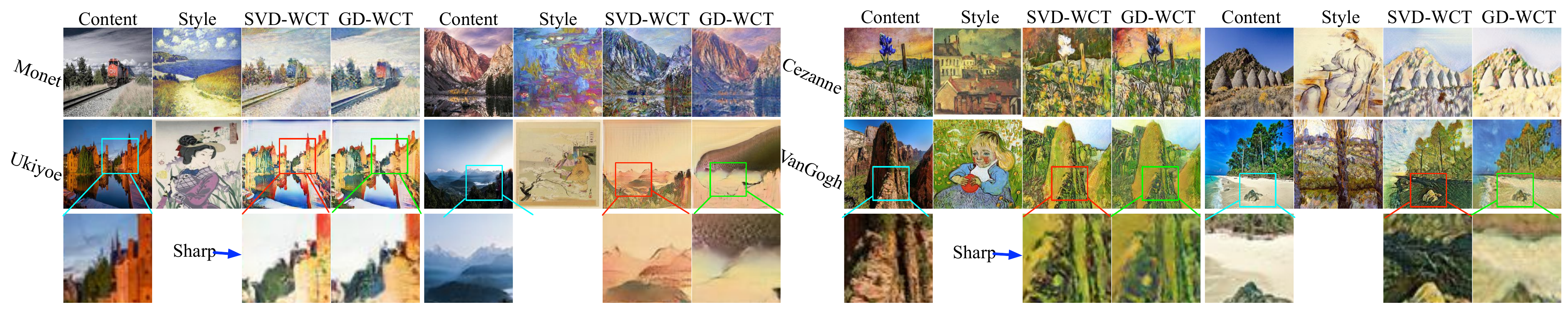}
	\vspace{-4mm}
	\caption{Qualitative comparisons on the Artworks dataset~\cite{Zhu17a}. Our method SVD-WCT generates images with sharper details than GD-WCT~\cite{Cho19}. Furthermore, the original style can be completely replaced by the new style. For instance, the colors of the train on the top left row and of the flower shown in the middle left row have changed completely. By contrast, with GD-WCT~\cite{Cho19}, the original style sometimes remains and the details are blurry. Better viewed with 8$\times$ zoom in. }
	\label{fig: gan}
\end{figure*}

\begin{table*}[t!]
	\caption{Classification accuracy (\%) of the real style images, and stylized images synthesized with the baseline GD-WCT~\cite{Cho19} and with our SVD-WCT. Our method consistently outperforms GD-WCT.}
	\label{tab: classification}
	\vspace{-2mm}
	\centering
	\begin{tabular}{lcccccccccccc}
		\toprule
		& \multicolumn{2}{c}{Cezanne} & \multicolumn{2}{c}{Monet} & \multicolumn{2}{c}{Real Photo} & \multicolumn{2}{c}{Ukiyoe} & \multicolumn{2}{c}{Vangogh} & \multicolumn{2}{c}{Mean Acc.} \\
		\midrule
		Model    & Acc.     & $\Delta$ Acc.    & Acc.    & $\Delta$ Acc.   & Acc.       & $\Delta$ Acc.     & Acc.     & $\Delta$ Acc.   & Acc.     & $\Delta$ Acc.    & Acc.      & $\Delta$ Acc.     
		\\ \midrule
		Original & 91.38    & -                & 97.52   & -               & 92.14      & -                 & 98.10    & -               & 100      & -                & 95.48     & -                 
		\\ \midrule
		GD-WCT   & 59.25    & 32.13            & 86.15   & 11.37           & 92.14      & 0                 & 49.13    & 48.97           & 36.88    & 63.12            & 64.71     & 30.77             \\
		Clip-WCT & 78.56    & 12.82            & 52.33   & 45.19           & 92.14      & 0                 & 49.67    & 48.43           & 28.63    & 71.37            & 60.27     & 35.21             \\
		PI-WCT   & 87.08    & \textbf{4.30}             & 78.16   & 19.36           & 92.14      & 0                 & 70.44    & \textbf{27.66}           & 69.64    & 30.36            & 79.49     & 15.99             \\
		SVD-WCT  & 85.89    & 5.49             & 96.27   & \textbf{1.25}            & 92.14      & 0                 & 68.04    & 30.06           & 74.17    & \textbf{25.83}            & 83.30     & \textbf{12.18}            
		\\ \bottomrule
	\end{tabular}
\end{table*}

\begin{figure*}
	\centering
	\includegraphics[width=\linewidth]{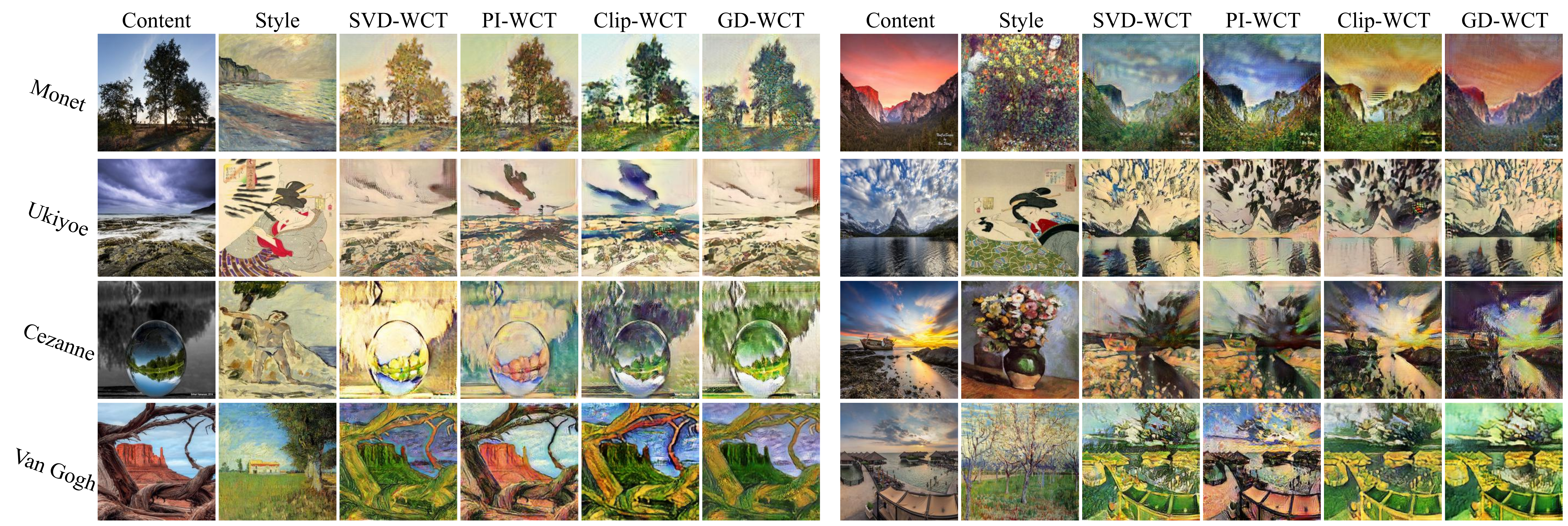}
	\vspace{-4mm}
	\caption{Comparisons between our SVD-WCT, the GD-WCT~\cite{Cho19}, the PI-WCT~\cite{Wang19b} and the Clip-WCT on the Artworks dataset~\cite{Zhu17a}.}
	\label{fig: wct_all}
	\vspace{-2mm}
\end{figure*}

\subsubsection{Network Structure}
Figure~\ref{fig: wct-pip} depicts the network used in~\cite{Cho19} to transfer the style of an image from domain $B$ to the content of an image from domain $A$. 
The main structure is an encoder-decoder network with 2 encoders: One content encoder $E_A^c$ used to model the content of image $x_A$ from domain $A$, and one style encoder $E_B^s$ aiming to represent the style of image $x_B$ from domain $B$. 
Multilayer perceptron networks (MLP) are used to compute coloring transformation matrices $(S_c^{(i)}, \mu_c^{(i)})$ for high-level and low-level features.
These matrices are then used to apply WCT consecutively to transfer the style from image $x_B$ to the content of image $x_A$, thus generating a stylized image $x_{A^{*}}$.
The resulting stylized image is passed to a discriminator whose goal is discriminate between real images from domain $B$ and generated images.
In the training process, the discriminator introduces an adversarial loss to help the generator to produce realistic images so as to fool the discriminator.
In the implementation of~\cite{Cho19}, another network transferring the style from domain $A$ to domain $B$ is trained in parallel, with an additional reconstruction loss encouraging the content to be preserved.

In essence, the process can be summarized as follows:
\begin{enumerate}[leftmargin=*]
	\item The content encoder $E_A^c$ encodes $x_A$ to $\mX_\mA$; 
	\item The style encoder $E_B^s$ encodes $x_B$ to $\mX_\mB$. Compute the coloring transformation matrices $(\mS_c^{(i)}, \mu_c^{i})$, $i=1,2,\cdots, 5$, from low-level to high-level features.
	\item Whiten $\mX_\mA$, color the resulting representation with $(\mS_c^{(i)}, \mu_c^{i})$, and pass the resulting features to convolutional layers to obtain a new $\mX_{\mA^*}$.
	\item Repeat (3) 5 times, followed by additional convolutions to generate the stylized image $x_{A^*}$.
\end{enumerate}

Because of its non-trivial backpropagation of this process, in~\cite{Cho19}, the WCT was learned by a small network using regularizers instead of SVD to obtain the whitening and coloring transformations. Here, we replace this with our differentiable SVD layer, which is non parametric. Since no regularizers are involved, there is no need to tune a regularizer weight. Thus, our approach is simpler to train.

We take the rest of the network and the remaining loss functions to be the same as in~\cite{Cho19}, which include the commonly-used adversarial loss~\cite{Goodfellow14b,Mao17,wang2020learning,wang2018every}, cycle loss~\cite{Zhu17a} and style loss~\cite{Lee18b,Huang18c}. For more detail, we refer the reader to~\cite{Cho19}.
We set the hyperparameter values in the SVD layer in the same manner as in our previous experiments, that is, $\epsilon{=}0.01$, and $K{=}9$. Training consumes about 21GPU hours on a single Titan XP (12GB). We name our Taylor based network SVD-WCT, and compare it to the method of~\cite{Cho19}, referred to as GD-WCT. As for image classification, we report the results of the Power Iteration method~\cite{Wang19b} (PI-WCT) and of gradient clipping (Clip-WCT) as baselines.
Note that we do not report the results of SVD-only and PI-only, because the large matrix size of the WCT task makes them typically fail.

\subsubsection{Performance}

We report results on the Artworks dataset~\cite{Zhu17a}, which contains 4 styles ({\ie} Monet, Ukiyoe, Cezanne, and Van Gogh).

\subsubsection*{Qualitative Analysis} 
In Figure~\ref{fig: gan}, we provide qualitative examples of stylized images. 
The content images are natural images taken in the wild, and the style images are randomly selected from the Artworks dataset. 
Note that our approach which computes a more accurate whitening matrix than~\cite{Cho19}, can better decorrelate the content and the style, thus better preserving the semantic content.
As a result, our SVD-WCT generates sharper images with a more detailed content than the baseline GD-WCT~\cite{Cho19}.  
This can be clearly seen, for instance, by looking at the building and mountain edges when transferring images to the Ukiyoe style (Row 2).
Similar observations can be made when transferring images to the Van Gogh style, where the rocks on the mountain and on the beach generated by our method  look sharper. 

Furthermore, our approach completely replaces the original colors by those from the reference style images.
For instance, the train (Row 1) and the flower (Row 4) are completely repainted with colors from the style image. 
By contrast, by using an approximate, less accurate whitening matrix, GD-WCT cannot disentangle the content and the style and thus preserves the original color of these objects.

\subsubsection*{Quantitative Analysis} 

To quantitatively evaluate our style transfer results, we rely on a classifier aiming to discriminate the different domains. 
The intuition behind this is that, for instance, an image transferred to the Monet style should be correctly classified as belonging to this style by a classifier pre-trained on real images from the four domains of the Artworks dataset. In essence, a higher accuracy of such a classifier indicates that the model better learned the patterns encoded in the target Monet style. 

In Table~\ref{tab: classification}, we compare the classification accuracies obtained for real images and images stylized with GD-WCT, PI-WCT, Clip-WCT and our SVD-WCT.
For this experiment, we used the pretrained Inception-v3~\cite{Szegedy16} and fine-tuned it on the training set of the Artworks dataset~\cite{Zhu17a}. 
Note that our SVD-WCT always yields higher accuracy than GD-WCT. Specifically, the gap between our accuracy and that obtained with the original Artworks test images is much smaller than that obtained using the GD-WCT images.
Furthermore, our SVD-WCT also outperforms Clip-WCT, which yields the worst performance because gradient clipping cannot preserve the descent direction. As shown in Figure~\ref{fig: wct_all}, in particular for the Monet and Cezanne styles,  Clip-WCT and GD-WCT cannot remove the style completely. PI-WCT yields the second-best performance. It is comparable to that of our SVD-WCT on all the styles except for the VanGogh one.

\begin{table}[t!]
	\caption{Sharpness scores. A larger score indicates a sharper image. Our SVD-WCT yields consistently larger values than GD-WCT~\cite{Cho19}.}
	\label{tab: gan}
	\vspace{-2mm}
	\centering
	\begin{tabular}{lcccc}
		\toprule
		& Cezanne & Monet & Ukiyoe & Vangogh \\ 
		\midrule
		GD-WCT~\cite{Cho19}  & 9.78   & 9.28  & 12.07   & 11.85    \\
		Clip-WCT & 11.65  &12.55  & 13.38  &  14.48   \\ 
		PI-WCT~\cite{Wang19b} & \textbf{11.98}  &\textbf{14.77}  & 14.68  &  16.56   \\ 
		SVD-WCT & 11.91    & 14.48  & \textbf{15.12}   & \textbf{18.67}    \\ \bottomrule
	\end{tabular}
\end{table}

As an additional quantitative metric, we compute a sharpness score~\cite{Sun08} for the stylized images. Specifically, we compute the sharpness of an image as the average norm of the intensity gradients along the vertical and horizontal image dimensions.
A larger value indicates a sharper image.
As shown in Table~\ref{tab: gan}, all the baselines using the standard SVD in the forward pass consistently outperform GD-WCT according to this metric. Our approach has similar performance to PI-WCT, outperforming gradient clipping. 

\begin{table}[t!]
	\caption{Reference time (ms) of different models.}
	\label{tab: ref-time}
	\vspace{-2mm}
	\centering
	\setlength\tabcolsep{4pt}
	\begin{tabular}{lcccc}
		\toprule
		Model          & GD-WCT~\cite{Cho19} & Clip-WCT & PI-WCT~\cite{Wang19b} & SVD-WCT \\
		\midrule
		Reference Time & 53     & 122      & 184    & 122   \\
		\bottomrule 
	\end{tabular}
	\vspace{2mm}
\end{table}

The average reference times of different models are shown in Table~\ref{tab: ref-time}. We report the average processing time for the entire test set. These timings were obtained with an NVIDIA GP102 Titan X GPU, and with batches of size 1. 
As shown in Figure~\ref{fig: wct-pip}, the channel of the input tensor has size 256. Following~\cite{Cho19}, we split the channels into 8 groups, and thus the input matrix to SVD has size 32.
Note that our new method, SVD-WCT, is much faster than our previous PI-based one, PI-WCT~\cite{Wang19b}. This is because PI-WCT involves a deflation process in the forward pass to process the eigenvectors sequentially, whereas our new approach processes the eigenvectors in parallel. The clipping method, \ie Clip-WCT has the same reference time as our SVD-WCT. This is because they have the same forward pass, both using a standard SVD. Note also that our SVD-WCT consumes more time than GD-WCT~\cite{Cho19}, but our method yields better results.

\section{Conclusion}
In this paper, we have introduced a Taylor-based approach to computing the SVD gradient. It is fast, accurate, and easy to incorporate in a deep learning framework. We have demonstrated the superiority of our approach over state-of-the-art ones for image classification and style-transfer.

This solves the backprogation issues that have long made the integration of eigendecomposition in deep networks problematic. In future work, to further increase the value of eigendecomposition for deep learning purposes, we will look into also speeding up the forward computation time.

\ifCLASSOPTIONcaptionsoff
  \newpage
\fi



\bibliographystyle{IEEEtran}
\bibliography{string,egbib}

\begin{thebibliography}{10}
\providecommand{\url}[1]{#1}
\csname url@samestyle\endcsname
\providecommand{\newblock}{\relax}
\providecommand{\bibinfo}[2]{#2}
\providecommand{\BIBentrySTDinterwordspacing}{\spaceskip=0pt\relax}
\providecommand{\BIBentryALTinterwordstretchfactor}{4}
\providecommand{\BIBentryALTinterwordspacing}{\spaceskip=\fontdimen2\font plus
\BIBentryALTinterwordstretchfactor\fontdimen3\font minus
  \fontdimen4\font\relax}
\providecommand{\BIBforeignlanguage}[2]{{%
\expandafter\ifx\csname l@#1\endcsname\relax
\typeout{** WARNING: IEEEtran.bst: No hyphenation pattern has been}%
\typeout{** loaded for the language `#1'. Using the pattern for}%
\typeout{** the default language instead.}%
\else
\language=\csname l@#1\endcsname
\fi
#2}}
\providecommand{\BIBdecl}{\relax}
\BIBdecl

\bibitem{Lei18}
L.~Huang, D.~Yang, B.~Lang, and J.~Deng, ``{Decorrelated Batch
  Normalization},'' in \emph{Conference on Computer Vision and Pattern
  Recognition}, 2018.

\bibitem{Li17a}
P.~Li, J.~Xie, Q.~Wang, and W.~Zuo, ``{Is Second-Order Information Helpful for
  Large-Scale Visual Recognition?}'' in \emph{International Conference on
  Computer Vision}, 2017.

\bibitem{Li18i}
P.~Li, J.~Xie, Q.~Wang, and Z.~Gao, ``{Towards Faster Training of Global
  Covariance Pooling Networks by Iterative Matrix Square Root Normalization},''
  in \emph{Conference on Computer Vision and Pattern Recognition}, 2018.

\bibitem{Huang19}
L.~Huang, Y.~Zhou, F.~Zhu, L.~Liu, and L.~Shao, ``{Iterative Normalization:
  Beyond Standardization Towards Efficient Whitening},'' in \emph{Conference on
  Computer Vision and Pattern Recognition}, 2019.

\bibitem{Yu17a}
K.~Yu and M.~Salzmann, ``{Second-Order Convolutional Neural Networks},'' in
  \emph{arXiv Preprint}, 2017.

\bibitem{Carreira12}
J.~Carreira, R.~Caseiro, J.~Batista, and C.~Sminchisescu, ``{Semantic
  Segmentation with Second-Order Pooling},'' in \emph{European Conference on
  Computer Vision}, 2012.

\bibitem{Ionescu15}
C.~Ionescu, O.~Vantzos, and C.~Sminchisescu, ``{Matrix Backpropagation for Deep
  Networks with Structured Layers},'' in \emph{Conference on Computer Vision
  and Pattern Recognition}, 2015.

\bibitem{Pan19a}
X.~Pan, X.~Zhan, J.~Shi, X.~Tang, and P.~Luo, ``{Switchable Whitening for Deep
  Representation Learning},'' in \emph{International Conference on Computer
  Vision}, 2019.

\bibitem{Chiu19}
T.-Y. Chiu, ``{Understanding Generalized Whitening and Coloring Transform for
  Universal Style Transfer},'' in \emph{International Conference on Computer
  Vision}, 2019.

\bibitem{Siarohin18}
A.~Siarohin, E.~Sangineto, and N.~Sebe, ``{Whitening and Coloring Batch
  Transform for GANs},'' in \emph{International Conference on Learning
  Representations}, 2018.

\bibitem{Miyato18}
T.~Miyato, T.~Kataoka, M.~Koyama, and Y.~Yoshida, ``{Spectral Normalization for
  Generative Adversarial Networks},'' in \emph{International Conference on
  Learning Representations}, 2018.

\bibitem{Li17d}
Y.~Li, C.~Fang, J.~Yang, Z.~Wang, X.~Lu, and M.-H. Yang, ``{Universal Style
  Transfer via Feature Transforms},'' in \emph{Advances in Neural Information
  Processing Systems}, 2017.

\bibitem{Cho19}
W.~Cho, S.~Choi, D.~K. Park, I.~Shin, and J.~Choo, ``{Image-To-Image
  Translation via Group-Wise Deep Whitening-And-Coloring Transformation},'' in
  \emph{Conference on Computer Vision and Pattern Recognition}, 2019.

\bibitem{Zanfir18c}
A.~Zanfir and C.~Sminchisescu, ``{Deep Learning of Graph Matching},'' in
  \emph{Conference on Computer Vision and Pattern Recognition}, 2018.

\bibitem{Dang18a}
Z.~Dang, K.~M. Yi, Y.~Hu, F.~Wang, P.~Fua, and M.~Salzmann,
  ``{Eigendecomposition-Free Training of Deep Networks with Zero
  Eigenvalue-Based Losses},'' in \emph{European Conference on Computer Vision},
  2018.

\bibitem{Yi18a}
K.~M. Yi, E.~Trulls, Y.~Ono, V.~Lepetit, M.~Salzmann, and P.~Fua, ``{Learning
  to Find Good Correspondences},'' in \emph{Conference on Computer Vision and
  Pattern Recognition}, 2018.

\bibitem{Lepetit09}
V.~Lepetit, F.~Moreno-Noguer, and P.~Fua, ``{EPnP: An Accurate O(n) Solution to
  the PnP Problem},'' \emph{International Journal of Computer Vision}, 2009.

\bibitem{Ferraz14}
L.~Ferraz, X.~Binefa, and F.~Moreno-Noguer, ``{Very Fast Solution to the {PnP}
  Problem with Algebraic Outlier Rejection},'' in \emph{Conference on Computer
  Vision and Pattern Recognition}, 2014, pp. 501--508.

\bibitem{Lewis96}
A.~S. Lewis, ``{Derivatives of Spectral Functions},'' \emph{Mathematics of
  Operations Research}, vol.~21, no.~3, pp. 576--588, 1996.

\bibitem{Nakatsukasa13}
Y.~Nakatsukasa and N.~J. Higham, ``{Stable and Efficient Spectral Divide and
  Conquer Algorithms for the Symmetric Eigenvalue Decomposition and the SVD},''
  \emph{SIAM Journal on Scientific Computing}, vol.~35, no.~3, pp. 1325--1349,
  2013.

\bibitem{Burden81}
R.~Burden and J.~Faires, \emph{Numerical Analysis}.\hskip 1em plus 0.5em minus
  0.4em\relax Cengage, 1989.

\bibitem{Bell97}
A.~J. Bell and T.~J. Sejnowski, ``{The “independent Components” of Natural
  Scenes Are Edge Filters},'' \emph{Vision research}, vol.~37, no.~23, pp.
  3327--3338, 1997.

\bibitem{Kessy18}
A.~Kessy, A.~Lewin, and K.~Strimmer, ``{Optimal Whitening and Decorrelation},''
  \emph{The American Statistician}, vol.~72, no.~4, pp. 309--314, 2018.

\bibitem{Wang19b}
W.~Wang, Z.~Dang, Y.~Hu, P.~Fua, and M.~Salzmann, ``{Backpropagation-Friendly
  Eigendecomposition},'' in \emph{Advances in Neural Information Processing
  Systems}, 2019.

\bibitem{Roy90}
R.~Roy, ``{The Discovery of the Series Formula for $\pi$ by Leibniz, Gregory
  and Nilakantha},'' \emph{Mathematics Magazine}, vol.~63, no.~5, pp. 291--306,
  1990.

\bibitem{Russakovsky15}
O.~Russakovsky, J.~Deng, H.~Su, J.~Krause, S.Satheesh, S.~Ma, Z.~Huang,
  A.~Karpathy, A.~Khosla, M.~Bernstein, A.~Berg, and L.~Fei-Fei, ``{Imagenet
  Large Scale Visual Recognition Challenge},'' \emph{International Journal of
  Computer Vision}, vol. 115, no.~3, pp. 211--252, 2015.

\bibitem{Papadopoulo00a}
T.~Papadopoulo and M.~Lourakis, ``{Estimating the Jacobian of the Singular
  Value Decomposition: Theory and Applications},'' in \emph{European Conference
  on Computer Vision}, 2000.

\bibitem{Rutishauser70}
H.~Rutishauser, ``{Simultaneous Iteration Method for Symmetric Matrices},''
  \emph{Numerische Mathematik}, vol.~16, no.~3, pp. 205--223, 1970.

\bibitem{Andrew14}
R.~Andrew and N.~Dingle, ``{Implementing QR Factorization Updating Algorithms
  on GPUs},'' \emph{Parallel Computing}, vol.~40, no.~7, pp. 161--172, 2014.

\bibitem{Liang19a}
J.~Liang, M.~Lin, and V.~Koltun, ``{Differentiable Cloth Simulation for Inverse
  Problems},'' in \emph{Advances in Neural Information Processing Systems},
  2019.

\bibitem{Lin15}
T.~Lin, A.~RoyChowdhury, and S.~Maji, ``{Bilinear Cnn Models for Fine-Grained
  Visual Recognition},'' in \emph{International Conference on Computer Vision},
  2015, pp. 1449--1457.

\bibitem{Lin17a}
T.~Y. Lin and S.~Maji, ``{Improved Bilinear Pooling with CNNs},'' in
  \emph{British Machine Vision Conference}, 2017.

\bibitem{Ioffe15}
S.~Ioffe and C.~Szegedy, ``{Batch Normalization: Accelerating Deep Network
  Training by Reducing Internal Covariate Shift},'' in \emph{International
  Conference on Machine Learning}, 2015.

\bibitem{Li17c}
Y.~Li, C.~Fang, J.~Yang, Z.~Wang, X.~Lu, and M.-H. Yang, ``{Diversified Texture
  Synthesis with Feed-Forward Networks},'' in \emph{Conference on Computer
  Vision and Pattern Recognition}, 2017.

\bibitem{Gatys16a}
L.~A. Gatys, A.~S. Ecker, and M.~Bethge, ``{Image Style Transfer Using
  Convolutional Neural Networks},'' in \emph{Conference on Computer Vision and
  Pattern Recognition}, 2016.

\bibitem{Gatys15}
L.~Gatys, A.~S. Ecker, and M.~Bethge, ``{Texture Synthesis Using Convolutional
  Neural Networks},'' in \emph{Advances in Neural Information Processing
  Systems}, 2015.

\bibitem{Mang17}
M.~Ye, A.~J. Ma, L.~Zheng, J.~Li, and P.~C. Yuen, ``{Dynamic Label Graph
  Matching for Unsupervised Video Re-Identification},'' in \emph{International
  Conference on Computer Vision}, 2017.

\bibitem{Huang20}
L.~Huang, L.~Zhao, Y.~Zhou, F.~Zhu, L.~Liu, and L.~Shao, ``{An Investigation
  into the Stochasticity of Batch Whitening},'' in \emph{Conference on Computer
  Vision and Pattern Recognition}, 2020.

\bibitem{He16a}
K.~He, X.~Zhang, S.~Ren, and J.~Sun, ``{Deep Residual Learning for Image
  Recognition},'' in \emph{Conference on Computer Vision and Pattern
  Recognition}, 2016, pp. 770--778.

\bibitem{Krizhevsky09}
A.~Krizhevsky, ``{Learning Multiple Layers of Features from Tiny Images},''
  Master's thesis, Department of Computer Science, University of Toronto, 2009.

\bibitem{TinyImageNet}
\BIBentryALTinterwordspacing
``{Tiny ImageNet}.'' [Online]. Available:
  \url{https://www.kaggle.com/c/tiny-imagenet}
\BIBentrySTDinterwordspacing

\bibitem{Krizhevsky12}
A.~Krizhevsky, I.~Sutskever, and G.~Hinton, ``{{ImageNet} Classification with
  Deep Convolutional Neural Networks},'' in \emph{Advances in Neural
  Information Processing Systems}, 2012, pp. 1106--1114.

\bibitem{Wang17b}
Q.~Wang, P.~Li, and L.~Zhang, ``G2denet: Global gaussian distribution embedding
  network and its application to visual recognition,'' in \emph{Conference on
  Computer Vision and Pattern Recognition}, 2017.

\bibitem{Chen17e}
D.~Chen, L.~Yuan, J.~Liao, N.~Yu, and G.~Hua, ``{Stylebank: An Explicit
  Representation for Neural Image Style Transfer},'' in \emph{Conference on
  Computer Vision and Pattern Recognition}, 2017.

\bibitem{Johnson16}
J.~Johnson, A.~Alahi, and L.~Fei{-}fei, ``{Perceptual Losses for Real-Time
  Style Transfer and Super-Resolution},'' in \emph{European Conference on
  Computer Vision}, 2016, pp. 694--711.

\bibitem{Ulyanov16b}
D.~Ulyanov, V.~Lebedev, A.~Vedaldi, and V.~S. Lempitsky, ``{Texture Networks:
  Feed-Forward Synthesis of Textures and Stylized Images},'' in
  \emph{International Conference on Machine Learning}, 2016.

\bibitem{Lee18b}
H.-Y. Lee, H.-Y. Tseng, J.-B. Huang, M.~Singh, and M.-H. Yang, ``{Diverse
  Image-To-Image Translation via Disentangled Representations},'' in
  \emph{European Conference on Computer Vision}, 2018.

\bibitem{Huang18c}
X.~Huang, M.-Y. Liu, S.~Belongie, and J.~Kautz, ``{Multimodal Unsupervised
  Image-To-Image Translation},'' in \emph{European Conference on Computer
  Vision}, 2018.

\bibitem{Huang17e}
X.~Huang and S.~Belongie, ``{Arbitrary Style Transfer in Real-Time with
  Adaptive Instance Normalization},'' in \emph{International Conference on
  Computer Vision}, 2017.

\bibitem{Zhu17a}
J.-Y. Zhu, T.~Park, P.~Isola, and A.~Efros, ``{Unpaired Image-To-Image
  Translation Using Cycle-Consistent Adversarial Networks},'' in
  \emph{International Conference on Computer Vision}, 2017, pp. 2223--2232.

\bibitem{Goodfellow14b}
I.~Goodfellow, J.~Pouget-Abadie, M.~Mirza, B.~Xu, D.~Warde-Farley, S.~Ozair,
  A.~Courville, and Y.~Bengio, ``{Generative Adversarial Nets},'' in
  \emph{Advances in Neural Information Processing Systems}, 2014.

\bibitem{Mao17}
X.~Mao, Q.~Li, H.~Xie, R.~Y. Lau, Z.~Wang, and S.~P. Smolley, ``{Least Squares
  Generative Adversarial Networks},'' in \emph{International Conference on
  Computer Vision}, 2017.

\bibitem{wang2020learning}
W.~Wang, X.~Alameda-Pineda, D.~Xu, E.~Ricci, and N.~Sebe, ``Learning how to
  smile: Expression video generation with conditional adversarial recurrent
  nets,'' \emph{IEEE Transactions on Multimedia}, vol.~22, no.~11, pp.
  2808--2819, 2020.

\bibitem{wang2018every}
W.~Wang, X.~Alameda-Pineda, D.~Xu, P.~Fua, E.~Ricci, and N.~Sebe, ``Every smile
  is unique: Landmark-guided diverse smile generation,'' in \emph{Conference on
  Computer Vision and Pattern Recognition}, 2018.

\bibitem{Szegedy16}
C.~Szegedy, V.~Vanhoucke, S.~Ioffe, J.~Shlens, and Z.~Wojna, ``{Rethinking the
  Inception Architecture for Computer Vision},'' in \emph{Conference on
  Computer Vision and Pattern Recognition}, 2016, pp. 2818--2826.

\bibitem{Sun08}
J.~Sun, Z.~Xu, and H.-Y. Shum, ``{Image Super-Resolution Using Gradient Profile
  Prior},'' in \emph{Conference on Computer Vision and Pattern Recognition},
  2008.

\end{thebibliography}
%
%
%

%


\begin{IEEEbiography}[{\includegraphics[width=1in,height=1.25in,clip,keepaspectratio]{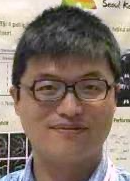}}]{Wei Wang} 
is an Assistant Professor of Computer Science at University of Trento, Italy.
Previously, after obtaining his PhD from University of Trento in 2018, he became a Postdoc at EPFL, Switzerland.
His research interests include machine learning and its application to computer vision and multimedia analysis.
\end{IEEEbiography}
\vspace{-25mm}

\begin{IEEEbiography}[{\includegraphics[width=1in,height=1.25in,clip,keepaspectratio]{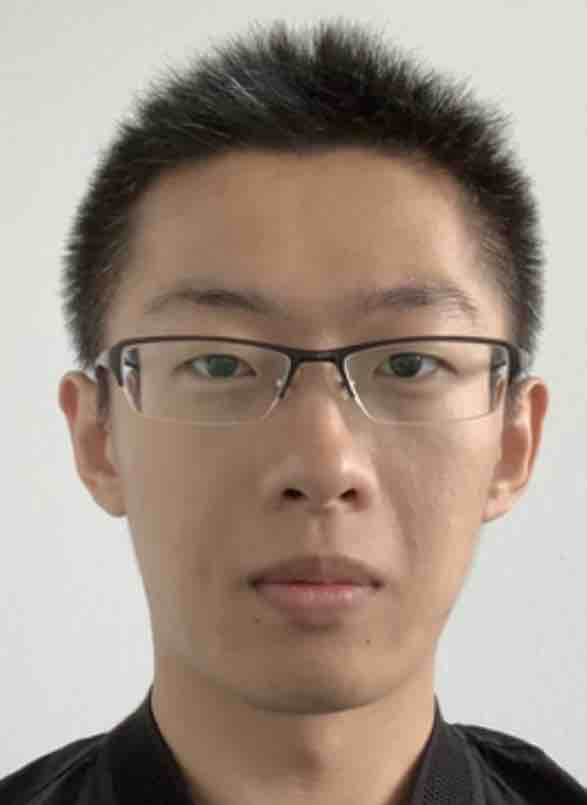}}]{Zheng Dang}
received the B.S. degree in Automation from Northwestern Polytechnical University in 2014. Currently he is a Ph.D. candidate and pursuing his Ph.D. degree in National Engineering Laboratory for Visual Information Processing and Application, Xian Jiaotong University, Xi’an, China. His research interests at how to bridge the gap between geometry and deep learning.
\end{IEEEbiography}
\vspace{-25mm}

\begin{IEEEbiography}[{\includegraphics[width=1in,height=1.25in,clip,keepaspectratio]{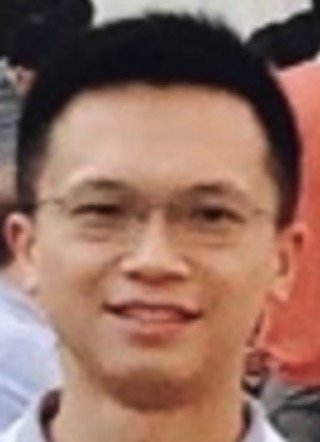}}]{Yinlin Hu}
received his M.S. and Ph.D. degrees in Communication and Information Systems from Xidian University, China, in 2011 and 2017 respectively. Before he started to pursue the Ph.D. degree, he was a senior algorithm
engineer and technical leader in Zienon, LLC. He is currently a post-doc in the CVLAB of EPFL. His research interests include optical flow, 6D pose estimation, and geometry-based learning methods.
\end{IEEEbiography}
\vspace{-25mm}

\begin{IEEEbiography}[{\includegraphics[width=1in,height=1.25in,clip,keepaspectratio]{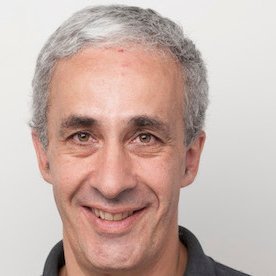}}]{Pascal Fua} 
is a Professor of Computer Science at EPFL, Switzerland. His research interests include shape and motion reconstruction from images, analysis of microscopy images, and Augmented Reality. He is an IEEE Fellow and has been an Associate Editor of the IEEE journal Transactions for Pattern Analysis and Machine Intelligence.
\end{IEEEbiography}
\vspace{-25mm}

\begin{IEEEbiography}[{\includegraphics[width=1in,height=1.25in,clip,keepaspectratio]{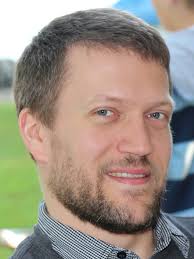}}]{Mathieu Salzmann}
is a Senior Researcher at EPFL and an Artificial Intelligence Engineer at ClearSpace. Previously, after obtaining his PhD from EPFL in 2009, he held different positions at NICTA in Australia, TTI-Chicago, and ICSI and EECS at UC Berkeley. His research interests lie at the intersection of machine learning and computer vision.
\end{IEEEbiography}

\newpage

\section{Appendix}

\begin{figure*}[t!]
	\centering
	\subfloat[Ratio $q_2 = \frac{\lambda_{2}}{\lambda_{1}} \leq 1$.]{\includegraphics[width=0.31\linewidth]{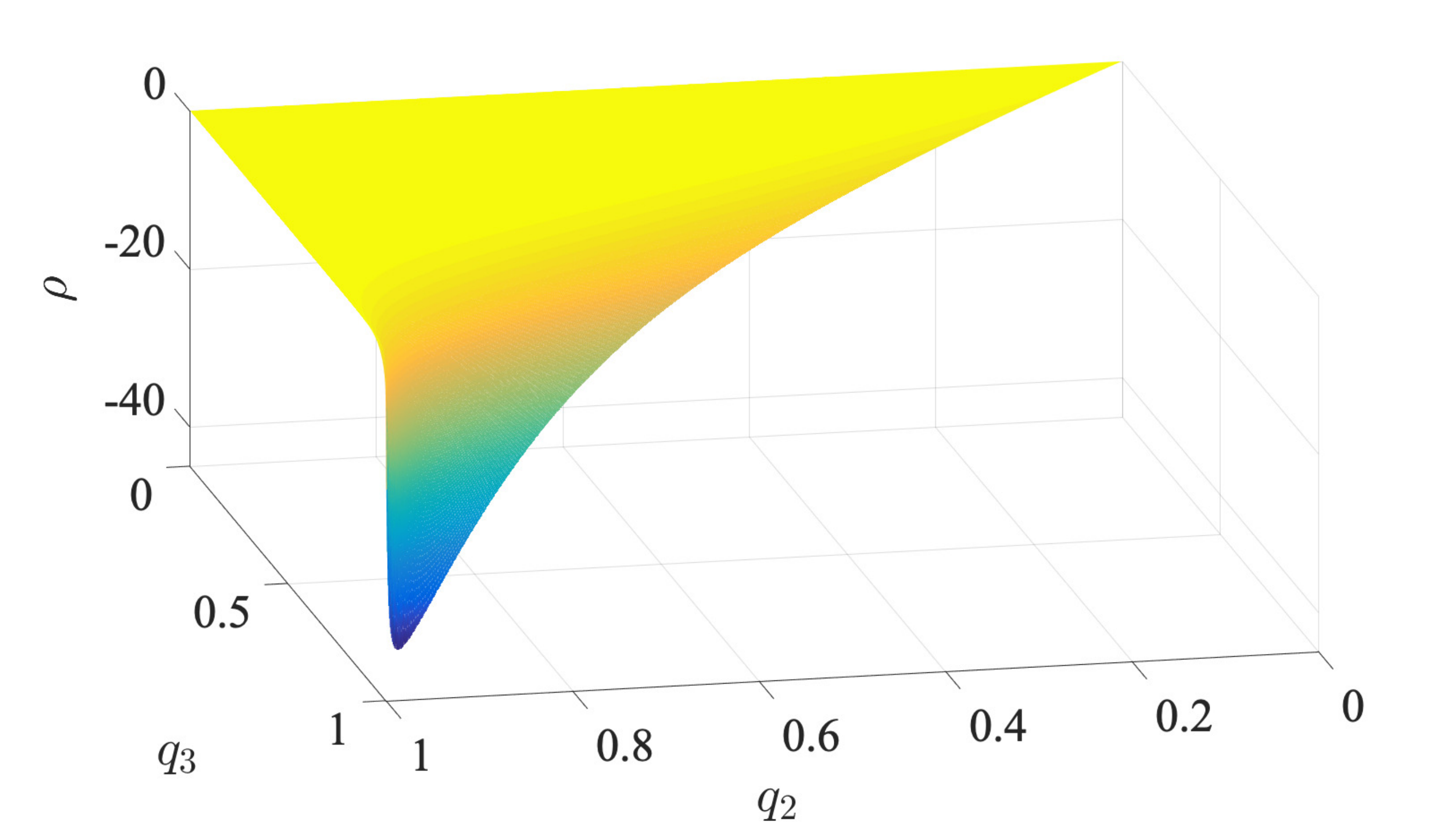} }%
	\quad
	\subfloat[Ratio $q_2 = \frac{\lambda_{2}}{\lambda_{1}} \leq 0.5$.]{\includegraphics[width=0.31\linewidth]{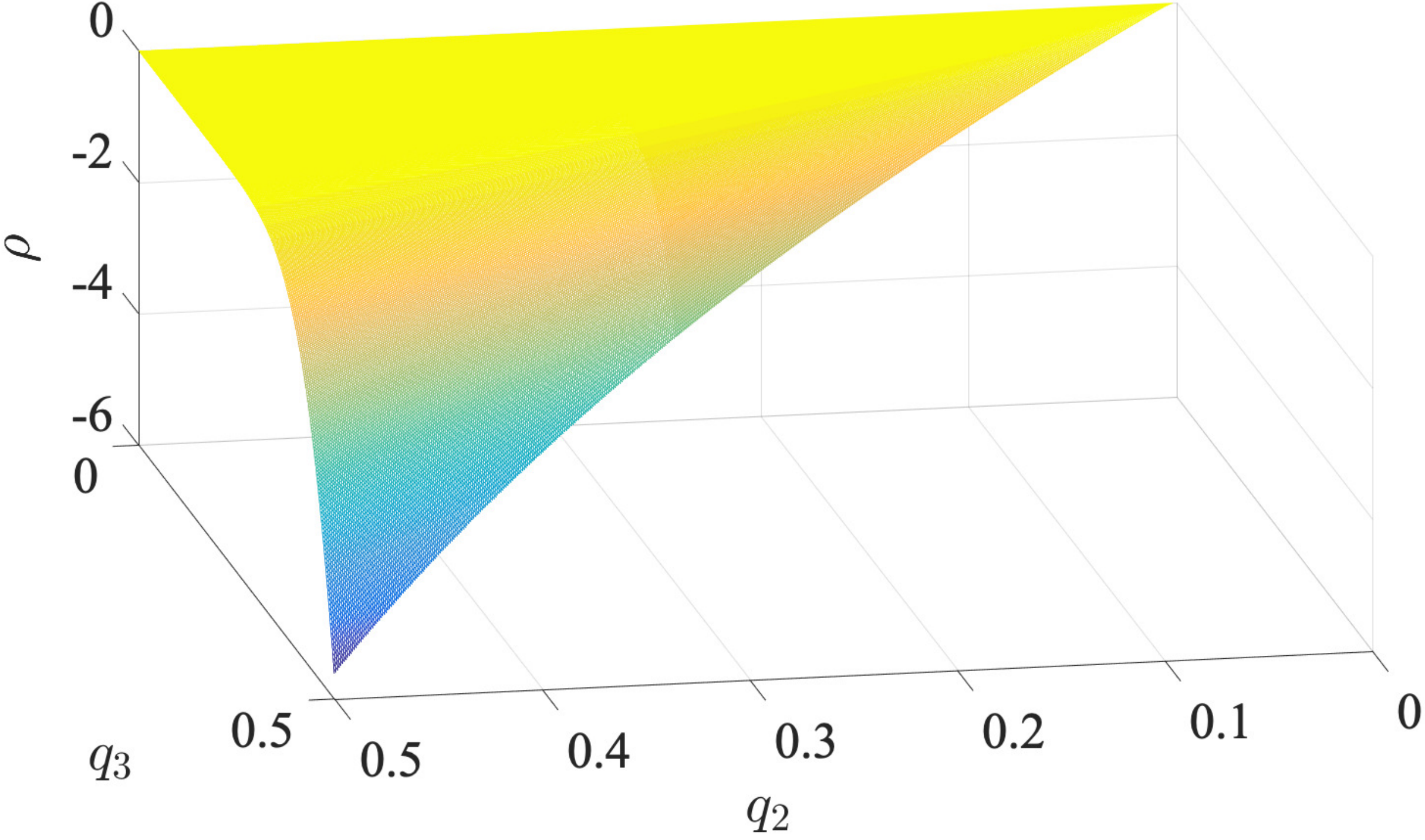} }%
	\quad
	\subfloat[Ratio $q_2 = \frac{\lambda_{2}}{\lambda_{1}} \leq 0.01$.]{\includegraphics[width=0.31\linewidth]{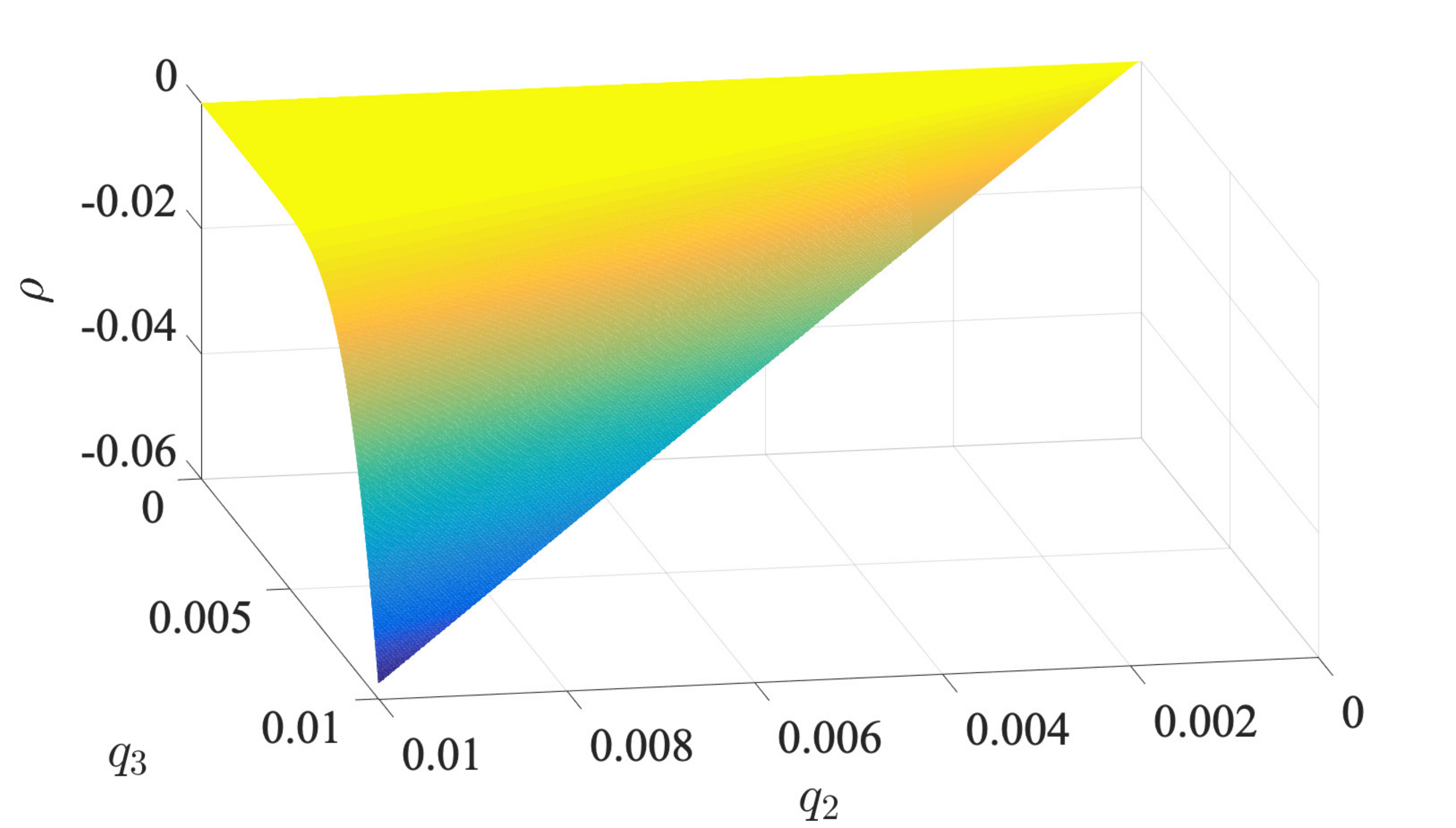} }%
	\caption{Value of $\rho {\in} [-90^{\circ}, 90^{\circ}]$ as a function of $q_2$ and $q_3$.}
	\label{fig: extreme}
\end{figure*}

\subsection{Bound of the Deviation from $\widetilde{K}$ to its Taylor form}
\label{sec: bound}
As shown in Eq.~\ref{eq: svd-cov}, the difference between the gradient computed using Taylor expansion and the original one only lies in the variable $\widetilde{\mK}$, which appears only once in the first term. The eigenvector matrix $\mV$ in Eq.~\ref{eq: svd-cov} is an orthogonal matrix in which the $l_2$ norm of each column vector $\vv_i$ is 1 such that the absolute value of the element in $\mV$ is always smaller than or equal to 1. The terms $\frac{\partial \mL}{\partial \mV}$ and $\frac{\partial \mL}{\partial \Lambda}_{diag}$ are the gradients backpropagated from the loss, which we observed to usually be very small. Therefore, the term which has the biggest influence on the value of Eq.~\ref{eq: svd-cov} is $\widetilde{\mK}$, whose elements become extremely large when two eigenvalues are close to each other.

To study the effect of the Taylor expansion on the gradient, we focus on the matrix $\widetilde{\mK}$ of Eq.~\ref{eq: unstable_term_eig}.
Below, for simplicity, we describe the scenario where $d=3$, that is, $\mM$ is a $3\times 3$ matrix, and we have 3 eigenvectors. Note, however, that our analysis extends to the general case by considering any hyperplane defined by 3 neighboring eigenvectors. In this scenario, $\widetilde{\mK}$ has the form
\begin{equation}
	\resizebox{0.88\linewidth}{!} 
	{$
		\left[\begin{array}{rrr}
			\widetilde{\mK}_{11} & \widetilde{\mK}_{12}  & \widetilde{\mK}_{13} \\
			\widetilde{\mK}_{21} & \widetilde{\mK}_{22}  & \widetilde{\mK}_{23} \\
			\widetilde{\mK}_{31} & \widetilde{\mK}_{32}  & \widetilde{\mK}_{33}
		\end{array}\right] {=}
		\left[\begin{array}{ccc}
			0 & \frac{1}{\lambda_{1}-\lambda_{2}}  & \frac{1}{\lambda_{1}-\lambda_{3}}\\
			\frac{1}{\lambda_{2}-\lambda_{1}} & 0  & \frac{1}{\lambda_{2}-\lambda_{3}} \\
			\frac{1}{\lambda_{3}-\lambda_{1}} & \frac{1}{\lambda_{3}-\lambda_{2}}  & 0
		\end{array}\right].
		$}
	\label{eq: K}
\end{equation}

As an example, let us now consider its $3$-rd column, \ie ($[\widetilde{\mK}_{13}, \widetilde{\mK}_{23}, 0]^{\top}$). Again, our analysis below easily extends to the other columns.  Since the third element is 0, we ignore it, and focus on the resulting 2D vector that lies in the canonical $xy$-plane. 
After Taylor expansion, we obtain the corresponding approximate 2D vector $[\widehat{\mK}_{13}, \widehat{\mK}_{23}]$. Specifically, we have
\begin{equation}
	\widehat{\mK}_{13}  = \frac{1}{\lambda_1}\left(1{+}\left(\frac{\lambda_3}{\lambda_1}\right){+}\cdots{+}\left(\frac{\lambda_3}{\lambda_1}\right)^{K} \right)\;,
	\label{eq: x1}
\end{equation}
which, if $\lambda_1 \neq \lambda_3$, can be re-written as
\begin{equation}
	\widehat{\mK}_{13}=\frac{1}{\lambda_{1}{-}\lambda_{3}} \left(1{-}\left(\frac{\lambda_3}{\lambda_1}\right)^{K{+}1}\right)\;.
	\label{eq: x1_}
\end{equation}
Similarly, we have
\begin{equation}
	\widehat{\mK}_{23} = \frac{1}{\lambda_2}\left(1{+}\left(\frac{\lambda_3}{\lambda_2}\right){+}\cdots{+}\left(\frac{\lambda_3}{\lambda_2}\right)^{K} \right)\;,
	\label{eq: x2}
\end{equation}
which, if $\lambda_2 \neq \lambda_3$, can be simplified as
\begin{equation}
	\widehat{\mK}_{23} = \frac{1}{\lambda_{2}{-}\lambda_{3}}\left(1{-}\left(\frac{\lambda_3}{\lambda_2}\right)^{K{+}1}\right)\;.
	\label{eq: x2_}
\end{equation}

To compare the true vector $[\widetilde{\mK}_{13}, \widetilde{\mK}_{23}]$ with the approximate one $[\widehat{\mK}_{13}, \widehat{\mK}_{23}]$, we observe the relative angle with the canonical $x$ direction, \ie the axis $[1,0]$. Let $\alpha$ be the angle between $[\widetilde{\mK}_{13}, \widetilde{\mK}_{23}]$ and this $x$ axis, and $\beta$ be the angle between $[\widehat{\mK}_{13}, \widehat{\mK}_{23}]$ and this $x$ axis. Then, we have
\begin{align}
	\tan(\alpha) & = \frac{\widetilde{\mK}_{23}}{\widetilde{\mK}_{13}} = \frac{\lambda_{1} - \lambda_{3}}{\lambda_{2}-\lambda_{3}}\;, \\
	\tan(\beta)  & = \frac{\widehat{\mK}_{23}}{\widehat{\mK}_{13}}= \frac{\lambda_{1} - \lambda_{3}}{\lambda_{2}-\lambda_{3}}\cdot \frac{1-\frac{\lambda_3}{\lambda_2}^{K{+}1}}{1-\frac{\lambda_3}{\lambda_1}^{K{+}1}}\;.
\end{align}
Since $\widetilde{\mK}_{13}{\geq}\widetilde{\mK}_{23}$, we have $\tan(\alpha) {\geq} 1$, and thus $\alpha {\in} (45^{\circ}, 90^{\circ})$. Similarly, we have $\beta {\in} (45^{\circ}, 90^{\circ})$.
The angular residual between the true direction $[\widetilde{\mK}_{13}, \widetilde{\mK}_{23}]$ and the approximate one $[\widehat{\mK}_{13}, \widehat{\mK}_{23}]$ is
$\rho = \beta {-} \alpha$, with $ \rho \in  (-45^{\circ}, 45^{\circ})$. Specifically,
\begin{equation}
	\resizebox{0.88\linewidth}{!} 
	{$
		\rho {=}
		\arctan\left(\frac{\lambda_{1} {-} \lambda_{3}}{\lambda_{2}{-}\lambda_{3}} {\cdot} \frac{1{-}\frac{\lambda_3}{\lambda_2}^{K{+}1}}{1{-}\frac{\lambda_3}{\lambda_1}^{K{+}1}} \right) {-}
		\arctan \left(\frac{\lambda_{1} {-} \lambda_{3}}{\lambda_{2}{-}\lambda_{3}} \right).
		$}
\end{equation}

To better understand this residual, let us consider a practical scenario where the Taylor expansion degree $K{=}9$. Furthermore, let $q_2 {=} \nicefrac{\lambda_{2}}{\lambda_{1}}{\in}[0,1]$, and $q_3 {=} \nicefrac{\lambda_{3}}{\lambda_{1}}{\in}[0,1]$, with $q_3 {\leq} q_2$. Then, we have
\begin{equation}
	\rho {=}
	\arctan\left(\frac{1 {-} q_{3}}{q_{2} {-} q_{3}}\cdot \frac{1 {-} \frac{q_3}{q_2}^{10}}{1 {-} q_3^{10}} \right) {-}
	\arctan\left(\frac{1 {-} q_{3}}{q_{2} {-} q_{3}} \right)\;.
\end{equation}
\comment{
	\begin{align}
		\frac{(q_3 - 1)}{(q_2 {-} q_3)^2 (\frac{(q_3 - 1)^2}{(q_2 - q_3)^2} {+} 1)} {-}
		\frac{    \frac{   (q_3^{10}/q_2^{10} {-} 1)(q_3 {-} 1) }{ (q_3^{10} - 1)(q_2 - q_3)^2 } {+} \frac{10q_3^{10}(q_3 - 1)} {q_2^{11}(q_3^{10} - 1)(q_2 - q_3) }  } { \frac{ (q_3^{10}/q_2^{10} - 1)^2(q_3 - 1)^2 } {(q_3^{10} - 1)^2(q_2 - q_3)^2} + 1}
	\end{align}
	
	\begin{align}
		\frac{  \frac{ q_3^{10}/q_2^{10} - 1 } { (q_3^{10} - 1)(q_2 - q_3)} {+} \frac{(q_3^{10}/q_2^{10} - 1)(q_3 - 1)} {(q_3^{10} - 1)(q_2 - q_3)^2} {+} \frac{10q_3^9(q_3 - 1)}  { q_2^{10}(q_3^{10} - 1)(q_2 - q_3) } {-}  \frac{ 10q_3^9(q_3^{10}/q_2^{10} - 1)(q_3 - 1) }  {(q_3^{10} - 1)^2(q_2 - q_3) }  } {  \frac{ (q_3^{10}/q_2^{10} - 1)^2(q_3 - 1)^2 } { (q_3^{10} - 1)^2(q_2 - q_3)^2 } + 1 } - \frac{ \frac{1}{q_2 - q_3} + \frac{q_3 - 1}{(q_2 - q_3)^2} }  { \frac{(q_3 - 1)^2}  {(q_2 - q_3)^2} + 1 }
	\end{align}
}
Figure~\ref{fig: extreme} (a) depicts the variation of $\rho$ as a function of $q_2$ and $q_3$.
Note that the minimum value, \ie large negative residual, is reached when $q_2\rightarrow 1$ and $q_3\rightarrow 1$.
Specifically, the minimum value and its corresponding position are
\begin{equation}
	\rho {\approx}{-}44.99^{\circ}\;,\; 
	q_3 {=} 1{-}1.0 {\times} 10^{-6}\;,\;
	q_2 {=} q_3 {+} 1.0 {\times} 10^{-10}\;.
\end{equation}
While this suggests that our approximation may entail a large angular residual, in practice, as observed in our image classification and style transfer experiments, the dominant eigenvalue is typically much larger than the other ones. That is, we virtually never observe $\nicefrac{\lambda_{2}}{\lambda_{1}} {\rightarrow} 1$. However, for a matrix of large dimension, one more often encounters the situation where two or more consecutive eigenvalues are similar to each other, as evidenced by Figure~\ref{fig:ratio-close-eigen}(a). 
Figure~\ref{fig: extreme} (a) shows the case where three eigenvalues are close to each other, but not strictly equal. If $\lambda_{1}$ is not dominant, and the eigenvalues are strictly equal \ie $\lambda_{1} {=} \lambda_{2} {=} \lambda_{3}$, we have

\begin{equation}
	[\widetilde{\mK}_{13}, \widetilde{\mK}_{23}] = \left[ \frac{1}{\lambda_1-\lambda_2}, \frac{1}{\lambda_2-\lambda_3} \right] = \left[ \infty, \infty \right]\;.
\end{equation}

\noindent Then we can obtain $\alpha {=} \arctan\left(\nicefrac{ \widetilde{\mK}_{23}}{ \widetilde{\mK}_{13}}\right) {=} 45^{\circ}$.
According to Eq.\ref{eq: x1} and Eq.~\ref{eq: x2}, we can obtain the approximate vector:

\begin{equation}
	\left[  \widehat{\mK}_{13},  \widehat{\mK}_{23} \right] = \left[ \frac{10}{\lambda_{1}},  \frac{10}{\lambda_{2}}\right]\;.
\end{equation}

\noindent Accordingly, we have 
\begin{equation}
	\beta = \arctan\left(\frac{ \widehat{\mK}_{23}}{ \widehat{\mK}_{13}}\right) = \arctan\left(1\right) = 45^{\circ}\;.
\end{equation}

The residual $\rho {=} \beta {-} \alpha {=} 0$. This means that when the eigenvalues are equal to each other, the Taylor expansion perfectly preserves the direction.

In Figure~\ref{fig: extreme} (b) \& (c), we visualize the much more common cases where $\nicefrac{\lambda_{2}}{\lambda_{1}} {\rightarrow} 0.5$, and $\nicefrac{\lambda_{2}}{\lambda_{1}} {\rightarrow} 0$.
In Figure~\ref{fig: extreme} (b), the minimum value is achieved when $q_2 {\rightarrow} 0.5$, and $q_3{\rightarrow} 0.5$. 
Let us consider the case where $q_3{=}q_2{=}0.5$. This indicates that $\lambda_{2}{=}\lambda_{3}$, and $\lambda_{1} {=} 2\lambda_{2}{=}2\lambda_{3}$. This, in turns, translates to
\begin{equation}
	[\widetilde{\mK}_{13}, \widetilde{\mK}_{23}] = \left[ \frac{1}{2\lambda_3-\lambda_3}, \frac{1}{\lambda_3-\lambda_3} \right] = \left[ \frac{1}{\lambda_3}, \infty \right]\;,
\end{equation}
which means that $\alpha = 90^{\circ}$. Furthermore, Eq.~\ref{eq: x1} and Eq.~\ref{eq: x2} become
\begin{equation}
	\left[  \widehat{\mK}_{13},  \widehat{\mK}_{23} \right] = \left[ \frac{1}{\lambda_{3}}\left(1{-}0.5^{10}\right),  \frac{10}{\lambda_{3}}\right]\;,
\end{equation}
which means that
\begin{equation}
	\beta = \arctan\left(\frac{ \widehat{\mK}_{23}}{ \widehat{\mK}_{13}}\right) = \arctan\left(\frac{10}{1-0.5^{10}}\right) \approx 84.29^{\circ}\;.
	\label{eq: beta_value}
\end{equation}
Therefore, the largest negative residual $\rho$ and its corresponding coordinates are
\begin{equation}
	\rho {=} \beta {-} \alpha {\approx} 84.29^{\circ} {-} 90^{\circ}  {=} {-}\mathbf{5.71}^{\circ}; \quad q_2{=}q_3{=}0.5.
\end{equation}
This means that, if the dominant eigenvalue covers at least 50\% of the energy, the angle difference is bounded by $5.71^{\circ}$.
In practice, however, $\lambda_{1}$ often accounts for much more than 50\% of the energy, in which case, as shown in Figure~\ref{fig: extreme} (b), the absolute value of the angle difference will be much smaller than $5.71^{\circ}$. 

As shown in Figure~\ref{fig: extreme} (c), for the case where $\frac{\lambda_{2}}{\lambda_{1}}{\leq} 0.01$, the minimum value is achieved when $q_2 {\rightarrow} 0.01$ and $q_3 {\rightarrow} 0.01$. Following the same reasoning as before, we have
\begin{equation}
	[\widetilde{\mK}_{13}, \widetilde{\mK}_{23}] = \left[ \frac{1}{100\lambda_3-\lambda_3}, \frac{1}{\lambda_3-\lambda_3} \right] = \left[ \frac{1}{99\lambda_3}, \infty \right]\;,
\end{equation}
which implies that $\alpha = 90^{\circ}$. In this case, however, our approximation becomes
\begin{equation}
	\left[  \widehat{\mK}_{13},  \widehat{\mK}_{23} \right] = \left[ \frac{1}{99\lambda_{3}}\left(1{-}0.01^{10}\right),  \frac{10}{\lambda_{3}}\right]\;,
\end{equation}
and thus
\begin{equation}
	\beta = \arctan(\frac{ \widehat{\mK}_{23}}{ \widehat{\mK}_{13}}) = \arctan(\frac{10 \times 99}{1-0.1^{10}}) \approx 89.94^{\circ}\;.
\end{equation}
Therefore, the largest residual $\rho$ and its corresponding coordinates are
\begin{equation}
	\rho {=} \beta {-} \alpha {\approx} 89.94^{\circ} {-} 90^{\circ}  {=} {-}\mathbf{0.06}^{\circ}\;, \quad q_2{=}q_3{=}0.01.
\end{equation}
In short, the more energy is covered by the dominant eigenvalue, the closer is the approximate column vector obtained using Taylor expansion to the true one in $\widetilde{\mK}$. As a result, the approximate gradient and the real one shown in Eq.~\ref{eq: svd-cov} are also closer. Otherwise, the deviation of the column vector in $\widetilde{\mK}$ will be propagated to the real gradient $\frac{\partial \mL}{\partial \mM}$.

\subsection{Limitation of Gradient Clipping Method}
\label{sec: grad-clip}
We first discuss the general scenario and then provide an example in Section~\ref{sec: example}. 
To study the general case for gradient clipping, we use the same setup as for Figure~\ref{fig: extreme} (b), \ie $\lambda_{2}{=}\lambda_{3}$, $\lambda_{1} {=} 2\lambda_{3}$. Let $t$ be a threshold value. Then, gradient clipping yields an approximate matrix $\mK$ whose two non-zero values in its third column are defined as
\begin{align}
	\left[ \widehat{\mK}_{13}, \widehat{\mK}_{23} \right] {=} \left[ \min\left(\frac{1}{\lambda_{3}}, t\right), t
	\right] {=}
	\begin{cases}
		\left[t, \quad t\right] & \text{if $t{\leq}\frac{1}{\lambda_{3}}$} \\
		\left[\frac{1}{\lambda_{3}}, t\right] & \text{if $t {>}\frac{1}{\lambda_{3}}$} \;.
	\end{cases}
\end{align}
This implies that the angle $\beta$ between the corresponding 2D vector and the canonical $x$ direction is such that
\begin{align}
	\tan(\beta) = \frac{\widehat{\mK}_{23}}{\widehat{\mK}_{13}}=
	\begin{cases}
		1 &\text{if $t{\leq}\frac{1}{\lambda_{3}}$} \\
		t\cdot \lambda_{3} & \text{if $t{>}\frac{1}{\lambda_{3}}$} 
	\end{cases}
\end{align}

\begin{align}
	\Longrightarrow\beta = 
	\begin{cases}
		45^{\circ} &\text{if $t {\leq} \frac{1}{\lambda_{3}}$} \\
		\arctan(t \cdot \lambda_{3})& \text{if $t{>}\frac{1}{\lambda_{3}}$} 
	\end{cases}
	\label{eq:clip_cases}
\end{align}
This shows that, in contrast with our Taylor-based approximation, the angle obtained using gradient clipping is not a fixed value. It depends on both the smallest eigenvalue $\lambda_{3}$ and the threshold $t$. It can be as large as $45^{\circ}$, \ie $\rho = \beta-\alpha = 45 - 90$. By contrast, the angle bound for our approach  is $5.71^{\circ}$.

To further put this in perspective of our Taylor-based approximation, we can use the result of Eq.~\ref{eq: beta_value} to separate the second case of Eq.~\ref{eq:clip_cases} into two subcases. This yields the following 3 scenarios:
\begin{align}
	\begin{cases}
		\beta = 45^{\circ} &\text{if $t {\leq} \nicefrac{1}{\lambda_{3}}$} \\
		45^{\circ} {<} \beta {<} 84.29^{\circ} & \text{if $\frac{1}{\lambda_{3}}{<} t {<}\frac{1}{\lambda_{3}}\frac{10}{1-0.5^{10}}$} \\
		\beta {\geq} 84.29^{\circ} & \text{if $t {\geq}\frac{1}{\lambda_{3}}\frac{10}{1-0.5^{10}}$} 
	\end{cases}
\end{align}
In turns, this means that
\begin{align}
	\begin{cases}
		\rho = -45^{\circ} &\text{a) if $t {\leq} \nicefrac{1}{\lambda_{3}}$} \\
		-45^{\circ} {<} \rho {<} {-}5.71^{\circ} & \text{b) if $\frac{1}{\lambda_{3}}{<} t {<}\frac{1}{\lambda_{3}}\frac{10}{1-0.5^{10}}$} \\
		\rho \leq -5.71^{\circ} & \text{c) if $t {\geq}\frac{1}{\lambda_{3}}\frac{10}{1-0.5^{10}}$} 
	\end{cases}
	\label{eq: rho}
\end{align}
When the threshold $t$ is fixed, there will always be an eigenvalue $\lambda_3$ that is small enough to satisfy either condition (a) or (b). As a consequence, the absolute value of the bound will always be larger than the one derived from Taylor expansion.

\subsubsection{Example}
\label{sec: example}
Let us now consider a concrete example where the clipping threshold $t=100$. 
To satisfy condition a) in Eq.~\ref{eq: rho}, and reflect the fact that two eigenvalues are often close if they are small, let $\lambda_{3}{=}0.01$, $\lambda_{2}{=}\lambda_{3}{=}0.01$, and $\lambda_{1}{=}2\lambda_{3}{=}0.02$.
After introducing these values in Eq.~\ref{eq: K}, we obtain
\begin{equation}
	\mK = \left[\begin{array}{ccc}
		0 & 100  & 100 \\
		-100 & 0  & \infty \\
		-100 & \infty  & 0
	\end{array}\right]\;,
\end{equation}
while its 9-th degree Taylor expansion is
\begin{equation}
	\widehat{\mK} = \left[\begin{array}{ccc}
		0 & 99.90  & 99.90 \\
		-99.90 & 0  & 1000 \\
		-99.90 & 1000 & 0
	\end{array}\right]\;.
\end{equation}
Even though the difference between $\widetilde{\mK}_{2,3}$ and $\widehat{\mK}_{2,3}$ is $\infty$, the angle between the true 3-rd column and its Taylor expansion ({\ie} $[100, \infty, 0]^{\top}$ and $[99.90, 1000, 0]^{\top}$) is only $5.71.^{\circ}$. In other words, the descent direction remains reasonably accurate, which greatly contributes to the stability of our method.
By contrast, if we clip the gradient to the predefined threshold 100, the angle between the true 3-rd column and the clipped one $[0, 100, 100]$ is $45^{\circ}$.

This is further illustrated by Figure~\ref{fig: clip}: Truncating the large value (\ie, $\widetilde{\mK}_{23}$) to $\widehat{\mK}_{23}$ by gradient clipping, makes the gradient lean towards the horizontal axis, changing the original direction $[\widetilde{\mK}_{13}, \widetilde{\mK}_{23}]$  ({\it black arrow}) to $[\widetilde{\mK}_{13}, \widehat{\mK}_{23}]$ ({\it blue arrow}).
If, instead, both the small and large values are modified, as with Taylor expansion, the approximate gradient direction remains closer to the original one, as indicated by the {\it red arrow} $[\widehat{\mK}_{13}, \widehat{\mK}_{23}]$.

In this particular example, the problem could of course be circumvented by using a larger threshold (\eg $10^4$), which may better preserve descent direction. However, in practice, one will always face smaller $\lambda_{3}$ values, \eg $10^{-4}$, which satisfy condition a) in Eq.~\ref{eq: rho}, and
whose truncated value will lead to descent direction errors as large as $45^{\circ}$.

\end{document}